\documentclass{article}

\usepackage{arxiv}
\usepackage{amsmath}
\usepackage[utf8]{inputenc} 
\usepackage[T1]{fontenc}    
\usepackage{hyperref}       
\usepackage{url}            
\usepackage{booktabs}       
\usepackage{amsfonts}       
\usepackage{nicefrac}       
\usepackage{microtype}      
\usepackage{lipsum}		
\usepackage{graphicx}
\usepackage{natbib}
\usepackage{doi}
\usepackage{tabularx}
\usepackage{color,soul}
\usepackage[table,dvipsnames]{xcolor}
\definecolor{lightblue}{rgb}{0.100, 0.700, 0.500}

\title{Rad-Phi2: Instruction Tuning Phi-2 for Radiology}

\author{ {Mercy Ranjit}\thanks{The author is also affiliated as a PhD student with Bharathidasan University.} \\
	Microsoft Research, India\\
	\texttt{meranjit@microsoft.com} \\
	\And
        {Gopinath Ganapathy} \\
	Department of Computer Science\\
	Bharathidasan University, India \\
	\texttt{gganapathy@bdu.ac.in} \\
	\AND
        {Shaury Srivastav} \\
    	Microsoft Research, India\\
    	\texttt{t-ssrivastav@microsoft.com} \\
        \And
        {Tanuja Ganu} \\
    	Microsoft Research, India\\
    	\texttt{taganu@microsoft.com} \\
        \And
        {Srujana Oruganti} \\
    	Microsoft Research, India\\
	\texttt{sruvs@microsoft.com} \\
}

\renewcommand{\shorttitle}{Rad-Phi2: Instruction Tuning Phi-2 for Radiology }

\hypersetup{
pdftitle={A template for the arxiv style},
pdfsubject={q-bio.NC, q-bio.QM},
pdfauthor={David S.~Hippocampus, Elias D.~Striatum},
pdfkeywords={First keyword, Second keyword, More},
}

\begin{document}
\maketitle

\begin{abstract}
Small Language Models (SLMs) have shown remarkable performance in general domain language understanding, reasoning and coding tasks, but their capabilities in the medical domain, particularly concerning radiology text, is less explored. In this study, we investigate the application of SLMs for general radiology knowledge specifically question answering related to understanding of symptoms, radiological appearances of findings, differential diagnosis, assessing prognosis, and suggesting treatments w.r.t diseases pertaining to different organ systems. Additionally, we explore the utility of SLMs in handling text-related tasks with respect to radiology reports within AI-driven radiology workflows.We fine-tune \texttt{Phi-2}, a SLM with 2.7 billion parameters using high-quality educational content from Radiopaedia \citep{radiopaedia2023radiopaedia}, a collaborative online radiology resource. The resulting language model, \texttt{RadPhi-2-Base}, exhibits the ability to address general radiology queries across various systems (e.g., chest, cardiac).
Furthermore, we investigate \texttt{Phi-2} for instruction tuning, enabling it to perform specific tasks. By fine-tuning \texttt{Phi-2} on both general domain tasks and radiology-specific tasks related to chest X-ray reports, we create \texttt{Rad-Phi2}. Our empirical results reveal that \texttt{Rad-Phi2 Base} and \texttt{Rad-Phi2} perform comparably or even outperform larger models such as \texttt{Mistral-7B-Instruct-v0.2} and \texttt{GPT-4} providing concise and precise answers.   In summary, our work demonstrates the feasibility and effectiveness of utilizing SLMs in radiology workflows both for knowledge related queries as well as for performing specific tasks related to radiology reports, thereby opening up new avenues for enhancing the quality and efficiency of radiology practice.
\end{abstract}


\section{Introduction}
Recent advances in natural language processing (NLP) have enabled the development of powerful language models that can generate natural language texts for various purposes and domains. However, most of the existing language models are trained on general domain texts, such as news articles, books, web pages, etc., and may not be able to handle the specific vocabulary, concepts, terminology, and style of radiology texts. Hence, it is required to have domain specific models that embody knowledge for that domain and trained for specific tasks in domain specific workflows to deal with the nuances of the field. But, the size and complexity of these language models are often prohibitive for practical applications, as they require a large amount of computational resources and memory to train and run. 

In this paper, we explore the use of small language models (SLMs) for radiology related tasks, specifically text-related tasks relating to radiology knowledge and tasks that are useful in everyday radiology workflows pertaining to radiology reports.

SLMs are language models that have a relatively small number of parameters, compared to the state-of-the-art large language models, such as \texttt{GPT-4} \citep{openai2023gpt4}. SLMs, specifically \texttt{Phi-2} \citep{gunasekar2023textbooks} is a 2.7 billion parameter model that has shown remarkable performance in general domain language understanding, reasoning, and coding tasks, but its capabilities in the medical domain, especially on radiology content and tasks specific to a radiology workflows is not explored. Moreover, \texttt{Phi-2} is not an instruction tuned model and tends to produce verbose answers as it was trained from synthetic text books and currently not suitable for direct deployments in application workflows to perform specific tasks. We hypothesize that SLMs can be effectively adapted to the radiology domain by finetuning them on high-quality radiology content to bring-in the nuanced radiology knowledge and also be adapted to perform specific tasks that are useful in everyday radiology workflows. SLMs provide the advantage of having a domain specific model for easy consumption at lower computational cost and faster inference. 

The \texttt{Phi} models \citep{gunasekar2023textbooks} success and its   on-par performance to large language models  was attributed to the high quality aka \textbf{\emph{“textbook-quality"}} training data that was used for its training. The training corpus was augmented with carefully selected data, filtered based on educational value and content quality. It demonstrated that this choice of training data enabled \texttt{Phi-2} to achieve outstanding performance on complex benchmarks, matching or outperforming models up to 25x larger in size. We extend this approach of \texttt{Phi-2} by using high-quality and reliable educational content from Radiopaedia \citep{radiopaedia2023radiopaedia} to finetune \texttt{Phi-2} for radiology knowledge. Radiopaedia is a rapidly growing collaborative online peer-reviewed open-edit radiology resource, compiled by radiologists and other health professionals from across the globe. Training from this highly valuable and reliable source can embody the model with this knowledge making it useful to answer questions on radiology accurately. To test our hypothesis, we use \texttt{Phi-2}, a 2.7 billion parameter SLM, as our base model, and finetune it on question-answer pairs derived from the educational articles of Radiopaedia pertaining to different systems like chest, cardiac, gastrointestinal etc. The resulting model \texttt{Rad-Phi2 Base}, can answer queries on radiology pertaining to these systems accurately. We evaluate the performance of \texttt{Rad-Phi2-Base} and compare it with larger instruction tuned models, such as \texttt{Mistral-7B-Instruct-v0.2} \citep{jiang2023mistral} which was finetuned on the same radiology question answer pairs and also \texttt{GPT-4}. \texttt{Rad-Phi2 Base} performs better than or comparable to the much larger counterparts while producing more crisp and to-the-point answers. It even surpassed the Retrieval Augmented Generation (RAG) approach \citep{lewis2021retrievalaugmented} that directly used a  knowledge base of Radiopaedia articles to answer the queries.

Apart from answering queries on radiology, there are specific text-related tasks that can be useful in AI deployment scenarios of radiology workflows. Some useful tasks relating to radiology reporting workflows are writing diagnostic impression summary from radiology findings, tagging of radiology reports with appropriate abnormality labels to flag priority cases, noise free extraction of findings and impression from existing radiology reports, generating a comparison summary of findings between the current radiology report and its prior report etc. Some of these tasks were also cited in the earlier work \citep{bhayana2024chatbots} in the context of application of language models in radiology workflows. To address these useful tasks that arise in the AI deployment scenarios of radiology workflows, we create an instruction tuning dataset for tasks pertaining to radiology reports of Chest X-rays and instruction-tune \texttt{Phi-2}  and obtain the \texttt{Rad-Phi2 model}. But, as \texttt{Phi-2} was not an instruction tuned model in the first place, we first instruction tune \texttt{Phi-2} for general domain tasks using the Super Natural instruction tuning dataset \citep{wang2022supernaturalinstructions}. The resulting model \texttt{Phi2-Instruct(SupNatInst)} is then further instruction tuned for tasks specific to radiology reports. The resulting model, \texttt{Rad-Phi2} can perform specific tasks that are useful in AI driven radiology reporting workflows of Chest X-rays. Our results show that \texttt{Rad-Phi2} performs better or on-par with a larger model like \texttt{Mistral-7B-Instruct(v0.2)} finetuned on the same instruction tuning dataset.

We would also like to note that while \texttt{GPT-4} has reasonable radiology knowledge as noted in this work \citep{liu2023exploring}, it still falls short on nuanced tasks like impression generation in comparison to domain specific models unless given with a large number of carefully curated examples in conjunction with sophisticated prompt engineering approaches\citep{Ma_2024}. SLMs enables to develop domain-specific models for these nuanced task types easily and quickly due to the low resource requirements associated with its size.

Our work demonstrates the feasibility and effectiveness of using SLMs for radiology specific tasks, and opens up new possibilities for improving the quality and efficiency of radiology practice. 

Our key Contributions include:

\begin{itemize}
\item We demonstrate the effectiveness of SLMs in the radiology domain by training on high quality radiology content from Radiopaedia \citep{radiopaedia2023radiopaedia}. The resulting model \texttt{Rad-Phi2-Base}  can answer general queries on radiology like symptoms, radiological appearances of findings, differential diagnosis, prognosis, and treatments w.r.t diseases pertaining to 15 different systems like Chest, Cardiac etc. It performs better than larger models like \texttt{Mistral-7B-Instruct-v0.2} finetuned on the same dataset. It also performed better than \texttt{GPT-4} and the Retrieval Augmented Generation approach that directly used Radiopaedia content as the knowledge source while answering queries.
\item We construct an instruction tuning dataset of  text related tasks that are useful in radiology report writing workflows of Chest X-rays using \texttt{GPT-4} processing and annotations of existing datasets on Chest X-ray reports. We share the prompts for different tasks used in creating this dataset.
\item We also demonstrate Instruction Tuning on SLMs to make it usable for practical tasks involved in radiology workflows. As \texttt{Phi-2} is not an instruction tuned model and tends to produce verbose answers, we first perform general domain instruction tuning using the Super Natural Instructions dataset followed by radiology specific instruction tuning on tasks related to radiology reports of Chest X-rays. We call the resulting model \texttt{Rad-Phi2} which is useful in performing  everyday tasks in radiology report writing workflows of Chest X-rays.
\item Our ablation studies gives insights on the input formats that worked the best when instruction tuning the \texttt{Phi-2} model. We note that radiology specific instruction tuning performs better when preceded by general domain instruction tuning and semantic tasks like Natural Language Inference(NLI) can help with the performance of other tasks.
\end{itemize}

\section{Related Work}
\label{sec:related_work}
The work, ‘Exploring the Boundaries of \texttt{GPT-4} in Radiology’ \citep{liu2023exploring} evaluated \texttt{GPT-4} on a diverse range of common radiology text-based tasks. It was noted that that zero-shot prompting technique was sufficient for the \emph{‘learn-by-instruction’} tasks like sentence similarity, radiology NLI, and disease progression. However, a large number of carefully curated examples with advanced prompt engineering were required for \emph{‘learn-by-example’} tasks like impression summary writing to perform on par with the supervised models trained for the task. While it is beneficial to have \texttt{GPT-4} like instruction-tuned models that can work well on radiology content for \emph{‘learn-by-instruction’} tasks, it is still beneficial to have specialized models for \emph{‘learn-by-example’} kind of tasks where the required knowledge for the model needs to be fed in as carefully curated in-context examples. It is even better if a model can do both these kinds of tasks with high accuracy at lower computational and resource costs, which is what we accomplish in our work.

The work ‘Chatbots and Large Language Models in Radiology: A Practical Primer for Clinical and Research Applications’\citep{bhayana2024chatbots} highlights how LLMs are useful in typical radiology workflows, providing key relevant information from clinical notes and prior reports, thus improving radiology interpretation and clinical decision making. The paper also noted LLMs when connected to sources like the American College of Radiology performed better than radiologists in applying this information to clinical scenarios. In regular practice, radiologists and clinicians refer to external knowledge sources like Radiopaedia \citep{radiopaedia2023radiopaedia} for information like clinical symptoms, radiological appearances of findings, differential diagnosis, prognosis, and treatment plan of a disease that aids them with clinical decision making. These knowledge sources are very reliable as they are maintained by practitioners of medical practice and are actively updated. However, this also poses a need for the AI models to be updated regularly with this knowledge. While Retrieval Augmented Generation \citep{lewis2021retrievalaugmented} can help LLMs augment their responses with the knowledge from the external sources, it may prove costly for high usage scenarios. This calls for a need for having Small Language Models that are easy to train, deploy, retrain, and iterate.

While instruction tuning to perform multiple tasks has been well explored on general domain tasks, it is less explored on radiology specific tasks. \texttt{Radiology-Llama2}\citep{liu2023radiologyllama2} and \texttt{Radiology-GPT}\citep{liu2023radiologygpt} performed instruction tuning for one specific task of generating clinically useful impression summary from radiological findings. However, there are a variety of tasks that can be useful in  radiology report writing workflows, like extracting a clean set of findings or impressions from a noisy radiology report, tagging the reports with abnormality labels or tubes, lines and devices labels, or comparing two reports to see the differential findings between them, querying the reports for specific questions or querying for general radiology knowledge with respect to various diseases like their radiographic appearances, symptoms, differential diagnosis or treatment plans to make a clinical decision. We consider all these useful tasks for instruction tuning. 

We explore the usage of Small Language Models for Radiology-specific tasks, as these models are very easy to train and deploy due to their smaller size, while being very competitive in performance to large language models. \texttt{Phi-2} paper \citep{gunasekar2023textbooks} introduced a 2.7 billion-parameter language model that demonstrated outstanding reasoning and language understanding capabilities, showcasing state-of-the-art performance among base language models with less than 13 billion parameters. On complex benchmarks, \texttt{Phi-2} matched or outperformed models up to 25x larger in size. The authors attributed this success to the high-quality textbook-like data they trained the model on. This is very applicable in the medical field, where the information rendered by the model needs to be accurate, which mandates the training dataset to be noise-free and of high quality. Our work considers this need of very high-quality data for training the models, which is practically deployable in day-to-day radiology workflows.

We also specifically note that \texttt{Phi-2} is a base model that has not undergone alignment through reinforcement learning from human feedback (RLHF) \citep{ouyang2022training}, nor has it been instruction finetuned. Due to this, it is noted that it often produces irrelevant or extra text following its first answer to user prompts within a single turn. This is due to its training dataset being primarily textbooks, which results in textbook-like responses. Our work addresses these issues by instruction tuning \texttt{Phi-2} first for general domain tasks to make it usable for downstream applications in Radiology.

In summary, our paper addresses the need for having special instruction tuned models for radiology which has accurate radiology knowledge and also can perform useful tasks on radiology workflows which is not fully addressed in the above stated works. In addition, we also perform these tasks using SLMs that are easy to train, deploy and iterate on without compromising on quality.

\section{Method}
\label{sec:Method}
\subsection{Dataset and Tasks}

\begin{table}
\centering
\caption{RADIOPAEDIA QUESTION ANSWERING DATA STATISTICS}
\begin{tabular}{l r r}
\hline
Systems & Article Counts & QA Pair Counts\\
\hline
Chest&1741&9695\\
Cardiac&771&5188\\
Central Nervous System&2846&16557\\
Urogenital&705&4171\\
Oncology&391&2782\\
Breast&357&1929\\
Musculoskeletal&3804&24026\\
Not Specified&1579&10880\\
Hepatobiliary&470&2956\\
Vascular&423&2304\\
Gastrointestinal&1240&7342\\
Obstetrics&568&3157\\
Interventional&149&1003\\
Trauma&85&483\\
Spine&94&490\\
Forensic&18&105\\
\hline
\multicolumn{1}{c}{Total}&15076&93068\\
\hline
\multicolumn{3}{p{240pt}}{Count of total articles and respective question answer pairs by system.}
\end{tabular}
\label{tab:Radiopaedia Stats}
\end{table}

\begin{table}
\centering
\caption{PROMPT DESIGN - INSTRUCTION TUNING DATASET FOR RADIOLOGY REPORTS}
\begin{tabular}{p{150pt} p{302pt}}
\hline
Task Name&Prompt\\
\hline
RADIOLOGY QA& \textless instruct\textgreater Question

\textless output\textgreater Answer\textless endoftext\textgreater\\
\hline
EXTRACT IMPRESSION&\textless instruct\textgreater Given the radiology report below:

FULL REPORT

Output the findings from the findings section of the radiology report. Each sentence in the output should describe an observation or a finding about the image. Do not mention any changes in observations, follow-up suggestions, doctor names, or noisy details. 

\textless output\textgreater Answer\textless endoftext\textgreater\\
\hline

EXTRACT FINDINGS&\textless instruct\textgreater Given the radiology report below:

FULL REPORT

Output the impression of the radiology report. Each sentence in the output should describe an observation or a finding about the image. Do not mention any changes in observations, follow-up suggestions, doctor names, or noisy details. 

\textless output\textgreater Answer\textless endoftext\textgreater\\
\hline

CLEANUP REPORT TEXT&\textless instruct\textgreater Given the text from a radiology report:

RADIOLOGY REPORT TEXT

Update the impressions or findings such that each sentence in the output describes an impression or observation about the image. Remove any mention of change of an observation and just state its presence. Do not include any follow-up suggestions or advice, and avoid mentioning any doctor names or other noisy details. 

\textless output\textgreater Answer\textless endoftext\textgreater\\
\hline
QA COMPREHENSION&\textless instruct\textgreater Answer the question using the radiology report below as context:

FULL REPORT

Question:

\textless output\textgreater Answer\textless endoftext\textgreater\\
\hline
QA TEMPORAL FINDINGS&\textless instruct\textgreater Given the below radiology report for an image and its prior report for reference: 

{FULL REPORT}

{PRIOR REPORT}

What findings are added and what findings are removed in the current radiology report for an image, compared to its reference report from before?
\textless output\textgreater Answer\textless endoftext\textgreater\\
\hline
QA TEMPORAL PROGRESSION&\textless instruct\textgreater Given the radiology report below, classify the progression of a finding as improved, no change, worsened.

{FULL REPORT}

\textless output\textgreater Answer\textless endoftext\textgreater\\
\hline
ABNORMALITY LABELS&\textless instruct\textgreater Given the below radiology report: 

{FULL REPORT}

What abnormality labels can be tagged to these findings?

\textless output\textgreater Answer\textless endoftext\textgreater\\
\hline
TUBES, LINES AND DEVICES LABELS&\textless instruct\textgreater Given the below radiology report:

FULL REPORT

Identify the tubes and lines or devices that are mentioned in the radiology report above. 

\textless output\textgreater Answer\textless endoftext\textgreater\\
\hline
IMPRESSION PREDICTION&\textless instruct\textgreater Given the findings from a radiology report:

{FINDINGS}

Based on the above findings from a radiology report, write an impression.

\textless output\textgreater Answer\textless endoftext\textgreater\\
\hline
\multicolumn{2}{p{400pt}}{Prompt Design for the Instruction Tuning dataset that was used to train \texttt{Rad-Phi2}.}
\end{tabular}
\label{tab:prompt design}
\end{table}

\begin{table}
\centering
\caption{RADIOLOGY REPORTS INSTRUCTION TUNING DATASET}
\begin{tabular}{l r r r r}
\hline
\multicolumn{1}{c}{Task} & Train & Test & Validate & Total\\
\hline
QA TEMPORAL PROGRESSION&50000&17639&7078&74717\\
ABNORMALITY LABELS&221035&3403&1959&226397\\
QA TEMPORAL FINDINGS&100000&2939&1308&104247\\
TUBES AND LINES DEVICES LABELS&94915&3403&1959&100277\\
IMPRESSION PREDICTION&208876&2523&1647&213046\\
EXTRACT FINDINGS&50000&3844&2103&55947\\
EXTRACT IMPRESSION&50000&3283&2295&55578\\
CLEANUP RADIOLOGY TEXT&75000&7127&4398&86525\\
QA COMPREHENSION&467057&9179&3878&480114\\
\hline
\multicolumn{1}{c}{Total} & 1316883	&53340	&26625	&1396848\\
\hline
\multicolumn{5}{p{340pt}}{Rad-Phi2 Instruction Tuning Dataset Statistics: Count of task wise splits for train, test and validation sets.}
\end{tabular}
\label{tab:radinstruct stats}
\end{table}

\paragraph{General Domain Instruction Tuning} We performed minimal general domain instruction tuning before we performed radiology specific finetuning using the Super Natural Instructions dataset \citep{wang2022supernaturalinstructions}. This dataset is a benchmark of 1,616 diverse NLP tasks including non-english tasks and their expert-written instructions. The collection covers 76 distinct task types. We used only the english specific tasks with a total of 757 tasks and 2,410,002 training records corresponding to these tasks.

\paragraph{Radiology Specific Instruction Tuning} We create two datasets for radiology specific finetuning, one for general radiology question answering and another for instruction tuning for tasks specific to radiology reports of Chest X-rays.

For radiology question answering task, we create question answer pairs using \texttt{GPT-4} from the articles of Radiopaedia pertaining to 15 different systems. We reserve the summary articles pertaining to these systems for testing. The count of question answer pairs w.r.t different systems is detailed in Table \ref{tab:Radiopaedia Stats}. This dataset was used to train the \texttt{Rad-Phi2-Base} model.

The instruction tuning dataset for tasks specific to radiology reports was constructed from the radiology reports of Mimic-CXR dataset \citep{Johnson2019-rs} that contains radiology reports pertaining to Chest X-rays. We use \texttt{GPT-4} processing and also annotations from the existing datasets derived from Mimic-CXR reports to construct this dataset. The prompts associated with the different tasks in the instruction tuning dataset is defined in Table \ref{tab:prompt design}. This dataset was used to train the \texttt{Rad-Phi2} model. The number of training records in the instruction tuning dataset for various tasks is detailed in Table \ref{tab:radinstruct stats}.

We construct the below tasks:

EXTRACT FINDINGS: Extract the findings from a given radiology report.

EXTRACT IMPRESSION: Extract the impression section from a given radiology report.

CLEANUP RADIOLOGY REPORT TEXT:
Given a radiology report text of findings or impression which includes noisy details like mention of Dr. names, followup actions etc., remove the noisy details to produce a clean radiology text of findings. We used \texttt{GPT-4} processing to generate a clean set of findings from a radiology report.

QA COMPREHENSION: 
Given a radiology report, answer the question based on the radiology report. The question could be about the abnormalities in the report, associated location or severity of findings etc. We use the Medical-Diff VQA dataset \citep{Hu2023-er} to construct the QA comprehension task.  

QA TEMPORAL FINDINGS:
Given a radiology report and its prior report, extract the differential set of findings between them. We use the Medical-Diff VQA dataset \citep{Hu2023-er} to construct the QA temporal comprehension task from the corresponding radiology reports.

QA TEMPORAL PROGRESSION:
Given a radiology report, classify the progression of an abnormality as improved, worsened or unchanged. We use the ChestImagenome dataset \citep{Wu2021-yi} for the progression labels.

ABNORMALITY LABELS: 
Given a radiology report, label the report as positive for a list of abnormalities. We tag the reports to findings/disease labels defined in the ChestImagenome dataset \citep{Wu2021-yi}. 

TUBES, LINES AND DEVICES LABELS:
Given a radiology report, label the report for the presence of tubes, line and devices. We tag the reports to tubes, lines and device labels defined in the ChestImagenome dataset \citep{Wu2021-yi}. 

IMPRESSION PREDICTION: 
Given the findings from a radiology report, summarize the findings as a diagnostic impression.

\subsection{Training Workflow}

\begin{figure}[h]
	\centering
     \includegraphics[width=\linewidth]{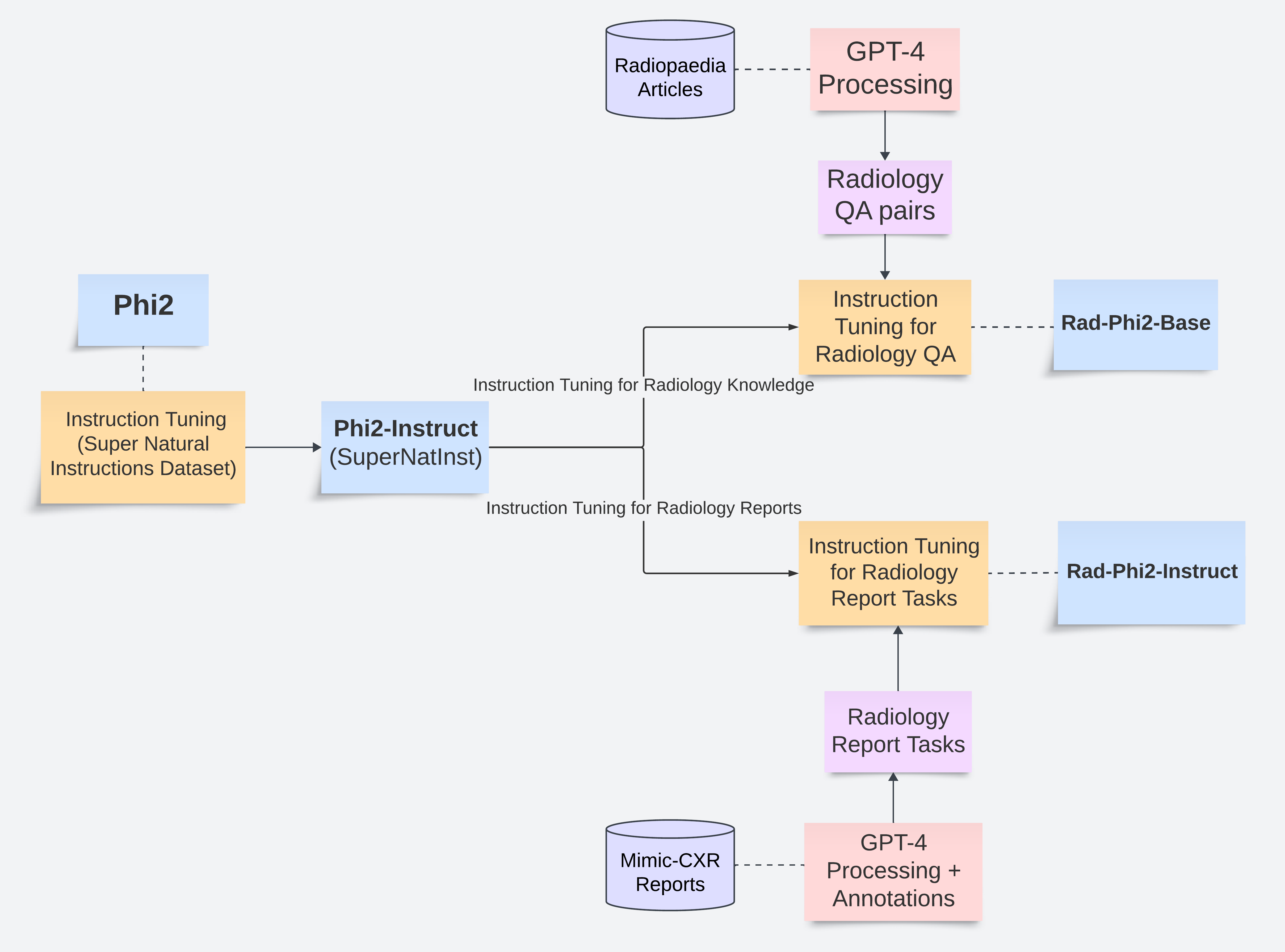}
	\caption{Rad-Phi2 Training Workflow}
	\label{fig:training_workflow}
\end{figure}

The training workflow is defined in the Fig.  \ref{fig:training_workflow}. We started with the \texttt{Phi-2} model and first instruction tuned it using the general domain tasks defined in the Super Natural Instructions dataset \citep{wang2022supernaturalinstructions}. We refer the resulting version as \texttt{Phi2-Instruct(SupNatInst)} . This was further finetuned for Radiology domain to produce two different models, \texttt{Rad-Phi2-Base} for answering radiology specific questions pertaining to 15 different systems defined in Table \ref{tab:Radiopaedia Stats} and \texttt{Rad-Phi2} for the tasks specific to radiology reports defined in Table \ref{tab:radinstruct stats}.

We use \texttt{GPT-4} for processing the articles from Radiopaedia \citep{radiopaedia2023radiopaedia} into question answer pairs for creating the Radiology QA dataset that was used to train the \texttt{Rad-Phi2-Base}. We also use it for processing the Radiology Reports to extract a clean set of findings from the radiology reports used in the  creation of the Extract Findings/Impression and Cleanup Radiology Text tasks of the instruction tuning dataset. We also use annotations from existing datasets and apply the prompts defined in Table \ref{tab:prompt design} with these annotations to create the instruction tuning dataset that was used in training \texttt{Rad-Phi2}.

\subsection{Experimental Design}

The experiments are grouped under three categories.
The first set of experiments are related to instruction tuning \texttt{Phi-2} model for general domain tasks. We do not intend to create a full-fledged instruction tuned \texttt{Phi-2} for general domain tasks in this work but our focus is to do minimal general domain instruction tuning as a preparation for radiology specific instruction tuning. We finetune   \texttt{Phi-2} using the Super Natural Instructions dataset. \citep{wang2022supernaturalinstructions}.   We evaluate the resulting model \texttt{Phi2-Instruct(SupNatInst)} on some unseen tasks that are defined in the test set of Super Natural Instructions  to test its generalization performance. We also measure the baseline performance of \texttt{Phi-2} on these sample set of tasks to check the improvements from the general domain instruction tuning. Table \ref{tab:experimentaldesign_supernli} documents the related experiments.

\begin{table}[hbt]
    \caption{INSTRUCTION TUNING EXPERIMENTS - SUPER NATURAL INSTRUCTIONS DATASET}
    
    \begin{tabular}{p{150pt} p{295pt}}
        \hline
        Model  & Description\\
        \hline
         Phi-2 Baseline& Baseline performance of \texttt{Phi-2} on general domain tasks.\\
        \hline
        Phi2-Instruct(SupNatInst)&Performance of \texttt{Phi-2}  model on general domain tasks after finetuning with Super Natural Instructions dataset. The resulting model is \texttt{Phi2-Instruct(SupNatInst)}\\
        \hline
        \multicolumn{2}{p{450pt}}{Experiments conducted w.r.t measuring the performance of Instruction Tuning  \texttt{Phi-2} with tasks defined in Super Natural Instructions dataset. The resulting model is \texttt{Phi2-Instruct(SupNatInst)}.} \\
    \end{tabular}
    \label{tab:experimentaldesign_supernli}
\end{table}

\begin{table}[hbt]
    \caption{RADIOLOGY QUESTION ANSWERING EXPERIMENTS}
    \begin{tabular}{p{150pt} p{295pt}}
        \hline
        Model & Description\\
        
        \hline
        Phi-2 Baseline& Baseline performance of \texttt{Phi2} on Radiology Question Answering without instruction tuning.\\
        \hline
        GPT-4 Baseline & Baseline performance \texttt{GPT-4} on Radiology Question Answering. We instruct \texttt{GPT-4} to answer in less than 50 words as otherwise it produced verbose answers.\\
        \hline
        RAG Baseline(\texttt{GPT-4}) & Performance of Retrieval Augmented Generation approach for Radiology Question Answering. Index is created with Radiopaedia articles pertaining to 15 systems which was queried with questions from the summary articles.\\ 
        \hline
        Rad-Phi2-Base&Performance on Radiology Question Answering after finetuning  \texttt{Phi2-Instruct(SupNatInst)} with QA pairs from Radiopaedia articles.\\
        \hline
        Mistral-7B-Instruct-v0.2&Performance on Radiology Question Answering after finetuning  \texttt{Mistral-7B-Instruct-v0.2} with QA pairs from Radiopaedia articles.\\
        \hline
        \multicolumn{2}{p{450pt}}{Experiments conducted with respect to measuring the performance of further Instruction Tuning \texttt{Phi-2-Instruct(SupNatInst)} for Radiology Question Answering. The resulting model is \texttt{Rad-Phi2 Base}.} \\
    \end{tabular}
    \label{tab:experimentaldesign_radiologyqa}
\end{table}

\begin{table}[hbt]
\centering

    \caption{INSTRUCTION TUNING EXPERIMENTS - RADIOLOGY REPORT TASKS}
    
    \begin{tabular}{p{150pt} p{295pt}} 
        \hline
        Model & Description\\
        \hline
        Phi-2 Baseline& Baseline performance of \texttt{Phi2} on tasks related to radiology reports.\\
        \hline
        GPT-4 Baseline& Baseline performance of \texttt{GPT-4} on tasks related to radiology reports. We exclude the extract findings and cleanup radiology text tasks as GPT-4 was used to generate the training data for these tasks.\\
        \hline
        Rad-Phi2&Performance after Instruction tuning \texttt{Phi2-Instruct(SupNatInst)} for radiology report related tasks. The resulting model is \texttt{Rad-Phi2}.\\
        \hline
        Mistral-7B-Instruct-v0.2&Performance after Instruction tuning \texttt{Mistral-7B-Instruct-v0.2} for radiology report related tasks.\\        
        \hline
        \multicolumn{2}{p{450pt}}{Experiments conducted with respect to measuring the performance of further Instruction Tuning \texttt{Phi2-Instruct(SupNatInst)} with tasks related to radiology reports. The resulting model is \texttt{Rad-Phi2}.} \\
    \end{tabular}
    \label{tab:experimental_design_report_tasks}
\end{table}

In the second set of experiments, we further finetune \texttt{Phi2-Instruct(SupNatInst)} for Radiology Question Answering. We consider three baseline experiments for this task - The baseline performance of \texttt{Phi-2}  for answering Radiology related questions, the baseline performance of \texttt{GPT-4} for answering Radiology related questions and Retrieval Augmented Generation(RAG) approach using \texttt{GPT-4} that uses an index of Radiology articles from Radiopaedia \citep{radiopaedia2023radiopaedia} to answer queries. We finetune \texttt{Phi-2} for Radiology Question Answering using QA pairs extracted from the Radiopaedia articles pertaining to 15 different systems defined in Table \ref{tab:Radiopaedia Stats}. We call this model \texttt{Rad-Phi2-Base}. We also finetune \texttt{Mistral-7B-Instruct-v0.2} using the same dataset for comparison. We reserve the summary articles of Radiopaedia for evaluation. Table \ref{tab:experimentaldesign_radiologyqa} documents the related experiments.

In the third set of experiments we finetune \texttt{Phi2-Instruct(SupNatInst)} for tasks that are useful in Radiology report writing workflows we defined in Table \ref{tab:radinstruct stats}. We also finetune \texttt{Mistral-7B-Instruct-v0.2} on the same dataset. We would like to note that \texttt{Mistral-7B-Instruct-v0.2} is an extensively instruction tuned model unlike \texttt{Phi-2} which was minimally finetuned only on Super Natural Instructions dataset to prepare it for Radiology specific tuning.  Table \ref{tab:experimental_design_report_tasks} documents the related experiments.
 
 \paragraph{Is GPT-4 sufficient for radiology workflows?} As we noted in the related work section
 [\ref{sec:related_work}], that while \texttt{GPT-4} perform extremely well on \textit{learn-by-instruction} tasks like sentence similarity, RadNLI, and disease progression classification, it required a large number of carefully curated examples for \textit{learn-by-example} tasks like impression summary writing to perform on par with the supervised models trained specifically for the task. We measure the baseline zero-shot performance of \texttt{GPT-4} for the \textit{learn-by-example} tasks like Impression prediction to elicit the need for  specialised models that are better suited for real radiology workflows.

\begin{table}
\parbox{.55\linewidth}{
\caption{COMPUTE REQUIREMENTS}
\begin{tabular}{p{60pt} p{25pt} p{25pt} p{30pt} p{40pt} }
\hline
Model Name&Train Time (GPU hrs)&Micro Batch (max)&Inference Speed (per token)&Inference Memory (2048 ctx)\\
\hline
Rad-Phi2-Base&60&8&2.78ms& \textless6.2G \\
\hline
Mistral-7B-Instruct-v0.2&800&2&3.33ms&\textless15.5G\\
\hline
\multicolumn{5}{p{240pt}}{Comparison of Compute using a single A100-80G with context length as 2048 and fp16 for the Radiology QA task.}
\end{tabular}
\label{tab:compute_reqs}
}
\hfill
\parbox{.43\linewidth}{

\caption{EXPERIMENTAL SETTINGS}
\begin{tabular}{p{85pt} p{80pt}}
\hline
Model Name&Parameters\\
\hline
Rad-Phi2-Base&Steps=19k,lr=$2e^{-5}$ \\
\hline
Mistral-7B-Instruct-v0.2(Radiology QA)& Steps=20k,lr=$2e^{-5}$ \\
\hline
Rad-Phi2& Steps=131k,lr=$2e^{-5}$ \\
\hline
Mistral-7B-Instruct-v0.2(Radiology Report Tasks)& Steps=133k,lr=$2e^{-5}$ \\
\hline
\multicolumn{2}{p{180pt}}{Adam optimizer was used in all the experiments.($\beta_1$=0.9$, \beta_2$=0.98$, \epsilon$=$1e^-7$) }
\end{tabular}
\label{tab:experimetal_settings}
}
\end{table}

The training and inference compute requirements for \texttt{Rad-Phi2} and \texttt{Mistral-7B-Instruct-v0.2} models are defined in Table \ref{tab:compute_reqs}. The experimental settings we used for training the models is defined in Table \ref{tab:experimetal_settings}.

\section{Results}
\subsection{Evaluation Metrics}
\paragraph{Lexical metrics}We use the lexical NLP metrics that measures the overlap between the predicted and reference texts. ROUGE-L \citep{lin2004rouge} measures the length of the longest common word sub sequence relative to the lengths of predicted and reference reports. BLEU-1 \citep{papineni-etal-2002-bleu} is based on uni-gram precision with a brevity penalty to discourage too short predictions. F1-score is a harmonic mean of unigram precision and recall. We use the implementation of \citep{li2023llavamed} for F1-score evaluations.

\paragraph{GPT-4 based Evaluation}As lexical metrics does not always necessarily provide an accurate evaluation for all cases, for example in a QA comprehension task where the model has to give an answer based on the context given, the predicted answer may be accurate in terms of semantics but the wording may be slightly different which results in a less accurate evaluation. To counter this, we use \texttt{GPT-4} based evaluation for QA comprehension tasks following the work of \citep{liu2023mitigating}, where \texttt{GPT-4} is used to come up with relevancy and accuracy scores to measure hallucinations in the predictions in a instruction tuning setting. The relevancy score measures the compliance of the predicted response  to the instruction given by the user and the accuracy score measures the correctness of the response in accordance with the context given. We pass the context, ground truth and predicted answers to \texttt{GPT-4} to measure the accuracy of the predictions. Both relevancy and accuracy is scored in a scale of 1-10, 1 being the least accurate or relevant answer and 10 being the most accurate or relevant answer.

\paragraph{Clinical Metrics}In addition to the above metrics, as we deal with radiology text generation for tasks like impression prediction, we also use a clinical metric named RadGraph F1 that measures the accuracy and completeness of the clinical facts generated by the  model. The RadGraph F1 is based on the RadGraph model \citep{jain2021radgraph} that parses the radiology reports into graphs containing clinical entities like anatomy and observations and relations between them. Entities are considered to match if the text spans and assigned types are the same, and relations are matched if their endpoint
entities and relation type are the same. We specifically use the Radgraph python package which is an implementation of the Radgraph F1 metric based on rewards \citep{jain2021radgraph}. The reward based scores are given based on the completeness of the extraction of entities and relationships using the RadGraph model. 

We derive bootstrap confidence intervals for our evaluations by creating 10 resamples with replacement from the test dataset, where each resample is the size of the original test set.

\begin{table}
    \centering 

     \caption{INSTRUCTION TUNING METRICS - SUPER NATURAL INSTRUCTIONS DATASET}
     \label{tab:supernli_metrics}

    \begin{tabularx}{\textwidth}{p{150pt} p{90pt} X X} 
        \hline
        Task &Metrics&Phi2-Baseline & Phi2-Instruct(SupNatInst) \\
        \hline
        task392\_inverse\_causal\_relationship
        &\begin{tabular}{l}F1-Score \\ Recall \\ Bleu-1 \\Rouge L\end{tabular}
        &\begin{tabular}{l}
        53.6 {\scriptsize \textcolor{gray}{ [52.56, 54.78]}}	\\ 
        50.82 {\scriptsize \textcolor{gray}{ [49.73, 52.0]}}	\\ 
        47.68 {\scriptsize \textcolor{gray}{ [46.67, 48.65]}}	\\ 
        53.6 {\scriptsize \textcolor{gray}{ [52.58, 54.75]}}\end{tabular}
        &
        \begin{tabular}{l}
        \textbf{59.96 {\scriptsize \textcolor{gray}{ [58.46, 61.39]}}}	\\ \textbf{56.32 {\scriptsize \textcolor{gray}{ [54.96, 57.64]}}}	\\ \textbf{52.02 {\scriptsize \textcolor{gray}{ [50.64, 53.36]}}}	\\ \textbf{59.94 {\scriptsize \textcolor{gray}{ [58.45, 61.43]}}}
        \end{tabular}\\
        \hline      
        
       task1529\_scitail1.1\_classification&
       \begin{tabular}{l}F1-Score \\ Recall \\ Bleu-1 \\ Rouge L\end{tabular}
        &
        \begin{tabular}{l}
        11.35 {\scriptsize \textcolor{gray}{ [10.46, 12.54]}} 
        \\ 11.28 {\scriptsize \textcolor{gray}{ [10.4, 12.46]}}	
        \\ 11.19 {\scriptsize \textcolor{gray}{ [10.32, 12.36]}}	
        \\ 11.45 {\scriptsize \textcolor{gray}{ [10.59, 12.64]}}\end{tabular}
        &
        \begin{tabular}{l}
        \textbf{44.84 {\scriptsize \textcolor{gray}{ [44.29, 45.76]}}} \\ \textbf{44.84 {\scriptsize \textcolor{gray}{ [44.29, 45.76]}}} \\ \textbf{44.84 {\scriptsize \textcolor{gray}{ [44.29, 45.76]}}} \\ \textbf{44.83 {\scriptsize \textcolor{gray}{ [44.26, 45.77]}}}
        \end{tabular}\\
        \hline

        task1615\_sick\_tclassify\_b\_relation\_a&
        \begin{tabular}{l}F1-Score \\ Recall \\ Bleu-1 \\ Rouge L\end{tabular}
        &
        \begin{tabular}{l}
        70.47 {\scriptsize \textcolor{gray}{ [68.74, 70.97]}} \\
        70.31 {\scriptsize \textcolor{gray}{ [68.59, 70.79]}}	 \\
        69.79 {\scriptsize \textcolor{gray}{ [68.03, 70.28]}}	\\ 
        69.93 {\scriptsize \textcolor{gray}{ [68.29, 70.45]}}\end{tabular}
        &
        \begin{tabular}{l}
        \textbf{77.99 {\scriptsize \textcolor{gray}{ [75.94, 78.37]}}} \\ \textbf{78.06 {\scriptsize \textcolor{gray}{ [76.0, 78.44]}}} \\ \textbf{77.71 {\scriptsize \textcolor{gray}{ [75.59, 78.06]}}} \\ \textbf{77.96 {\scriptsize \textcolor{gray}{ [75.92, 78.31]}}}
        \end{tabular}\\
        \hline

        task202\_mnli\_contradiction\_classification&
        \begin{tabular}{l}F1-Score \\ Recall \\ Bleu-1 \\ Rouge L\end{tabular}
        &
        \begin{tabular}{l}
        2.63 {\scriptsize \textcolor{gray}{ [2.56, 2.88]}}	\\ 
        1.62 {\scriptsize \textcolor{gray}{ [1.58, 1.84]}}	\\ 
        0.38 {\scriptsize \textcolor{gray}{ [0.33, 0.55]}}	\\ 
        2.62 {\scriptsize \textcolor{gray}{ [2.55, 2.87]}}\end{tabular}
        &
        \begin{tabular}{l}
        \textbf{67.53 {\scriptsize \textcolor{gray}{ [66.14, 68.48]}}} \\ \textbf{67.44 {\scriptsize \textcolor{gray}{ [66.05, 68.4]}}} \\ \textbf{67.35 {\scriptsize \textcolor{gray}{ [65.97, 68.31]}}} \\ \textbf{67.7 {\scriptsize \textcolor{gray}{ [66.23, 68.64]}}}
        \end{tabular}\\
        \hline

        task200\_mnli\_entailment\_classification
        &
        \begin{tabular}{l}F1-Score \\ Recall \\  Bleu-1 \\ Rouge L\end{tabular}
        &
        \begin{tabular}{l}
        3.87 {\scriptsize \textcolor{gray}{ [3.78, 4.03]}}	\\
        2.38 {\scriptsize \textcolor{gray}{ [2.32, 2.5]}}	\\
        0.59 {\scriptsize \textcolor{gray}{ [0.5, 0.74]}}	\\
        3.8 {\scriptsize \textcolor{gray}{ [3.73, 3.98]}}
        \end{tabular}
        &
        \begin{tabular}{l}
        \textbf{49.97 {\scriptsize \textcolor{gray}{ [48.93, 50.68]}}}\\
        \textbf{49.7 {\scriptsize \textcolor{gray}{ [48.65, 50.41]}}}\\
        \textbf{49.47 {\scriptsize \textcolor{gray}{ [48.39, 50.17]}}}\\
        \textbf{49.95 {\scriptsize \textcolor{gray}{ [48.95, 50.67]}}}\end{tabular}\\
        \hline

        task050\_multirc\_answerability&
        \begin{tabular}{l}F1-Score \\ Recall \\ Bleu-1 \\ Rouge L\end{tabular}
        &
        \begin{tabular}{l}
        41.35 {\scriptsize \textcolor{gray}{ [40.08, 41.82]}}	\\
        40.79 {\scriptsize \textcolor{gray}{ [39.53, 41.27]}}	\\
        40.3 {\scriptsize \textcolor{gray}{ [39.06, 40.8]}}	\\
        41.33 {\scriptsize \textcolor{gray}{ [40.05, 41.83]}}
        \end{tabular}
        &
        \begin{tabular}{l}
        \textbf{68.47 {\scriptsize \textcolor{gray}{ [68.35, 69.63]}}} \\ \textbf{68.46 {\scriptsize \textcolor{gray}{ [68.35, 69.62]}}} \\ \textbf{68.45 {\scriptsize \textcolor{gray}{ [68.35, 69.61]}}} \\ \textbf{68.47 {\scriptsize \textcolor{gray}{ [68.36, 69.67]}}}
        \end{tabular}\\
        \hline
        
        task827\_copa\_commonsense\_reasoning&
        \begin{tabular}{l}F1-Score \\ Recall \\ Bleu-1 \\ Rouge L\end{tabular}
        &
        \begin{tabular}{l}
        \textbf{85.11 {\scriptsize \textcolor{gray}{ [83.93, 86.66]}}} \\ \textbf{84.99 {\scriptsize \textcolor{gray}{ [83.85, 86.58]}}} \\ \textbf{84.9 {\scriptsize \textcolor{gray}{ [83.78, 86.51]}}} \\ \textbf{85.21 {\scriptsize \textcolor{gray}{ [83.94, 86.71]}}}
        \end{tabular}
        &
        \begin{tabular}{l}
        82.8 {\scriptsize \textcolor{gray}{ [79.69, 85.15]}}	\\
        82.78 {\scriptsize \textcolor{gray}{ [79.68, 85.13]}}	\\
        82.77 {\scriptsize \textcolor{gray}{ [79.68, 85.11]}}	\\ 
        82.83 {\scriptsize \textcolor{gray}{ [79.77, 85.13]}}\end{tabular}\\
        \hline
        
        task1390\_wscfixed\_coreference&
        \begin{tabular}{l}F1-Score \\ Recall \\ Bleu-1 \\ Rouge L\end{tabular}
        &
        \begin{tabular}{l}
        44.51 {\scriptsize \textcolor{gray}{ [42.28, 46.81]}}	\\ 
        44.37 {\scriptsize \textcolor{gray}{ [42.19, 46.72]}}	\\ 
        44.27 {\scriptsize \textcolor{gray}{ [42.12, 46.65]}}	\\ 
        44.5 {\scriptsize \textcolor{gray}{ [42.32, 46.8]}}\end{tabular}
        &
        \begin{tabular}{l}
        \textbf{51.39 {\scriptsize \textcolor{gray}{ [48.01, 53.58]}}} \\ \textbf{51.39 {\scriptsize \textcolor{gray}{ [48.01, 53.58]}}} \\ \textbf{51.39 {\scriptsize \textcolor{gray}{ [48.01, 53.58]}}} \\ \textbf{51.54 {\scriptsize \textcolor{gray}{ [47.98, 53.49]}}}
        \end{tabular}\\
        \hline
        
        task020\_mctaco\_span\_based\_question
        &
        \begin{tabular}{l}F1-Score \\ Recall \\ Bleu-1 \\ Rouge L\end{tabular}
        &
        \begin{tabular}{l}
        14.09 {\scriptsize \textcolor{gray}{ [11.54, 16.69]}}	\\ 
        13.89 {\scriptsize \textcolor{gray}{ [11.35, 16.52]}}	\\ 
        13.71 {\scriptsize \textcolor{gray}{ [11.15, 16.36]}}	\\ 
        15.0 {\scriptsize \textcolor{gray}{ [12.65, 17.21]}}\end{tabular}
        &
        \begin{tabular}{l}
        \textbf{31.16 {\scriptsize \textcolor{gray}{ [26.96, 37.71]}}}	\\ \textbf{30.6 {\scriptsize \textcolor{gray}{ [26.62, 36.83]}}} \\ \textbf{30.06 {\scriptsize \textcolor{gray}{ [26.25, 36.0]}}} \\ \textbf{31.21 {\scriptsize \textcolor{gray}{ [26.96, 37.54]}}}
        \end{tabular}\\
        \hline
        \multicolumn{4}{p{450pt}}{Instruction tuning with Super Natural Instructions dataset improves the performance of the sample tasks in-comparison to the Phi2-Baseline except for the copa\_commonsense task. Bootstrap confidence intervals for our evaluations are measured by creating 10 resamples with replacement from the test dataset, where each resample is the size of the original test set.}\\
        
\end{tabularx} 
\end{table}

\begin{table}
    \caption{RAD-PHI2-BASE - RADIOLOGY QA METRICS (ALL SYSTEMS)}
    \label{tab:metrics_radiopaedia}
    
    \begin{tabularx}{\textwidth}{p{110pt} X X X X} 
        \hline
        
        Model & F1-Score & Recall & BLEU-1 &  RougeL\\
        \hline
        Phi2-Baseline & 15.08 {\scriptsize \textcolor{gray}{ [13.18, 16.95]}}& 10.27 {\scriptsize \textcolor{gray}{ [8.78, 11.78]}} & 2.02 {\scriptsize \textcolor{gray}{ [1.38, 2.88]}} & 9.46 {\scriptsize \textcolor{gray}{ [8.48, 10.48]}}\\
        \hline
        Rad-Phi2-Base & \textbf{34.86 {\scriptsize \textcolor{gray}{ [30.26, 39.59]}}} & \textbf{39.48 {\scriptsize \textcolor{gray}{ [33.05, 45.38]}}} & \textbf{22.84 {\scriptsize \textcolor{gray}{ [18.54, 27.47]}}} &  \textbf{25.13 {\scriptsize \textcolor{gray}{ [21.3, 29.24]}}} \\
        \hline
        Mistral-7B-Instruct-v0.2 & 29.4 {\scriptsize \textcolor{gray}{ [25.43, 33.19]}} & 35.53 {\scriptsize \textcolor{gray}{ [29.78, 40.39]}} & 17.76 {\scriptsize \textcolor{gray}{ [14.59, 20.89]}} & 20.88 {\scriptsize \textcolor{gray}{ [18.04, 23.88]}} \\
        \hline
        GPT-4 & 31.54 {\scriptsize \textcolor{gray}{ [28.36, 33.87]}} & 28.28 {\scriptsize \textcolor{gray}{ [24.7, 30.62]}} & 18.52 {\scriptsize \textcolor{gray}{ [15.61, 21.07]}} &  21.64 {\scriptsize \textcolor{gray}{ [19.2, 23.6]}}\\
        \hline
        RAG(GPT-4) & 29.93 {\scriptsize \textcolor{gray}{ [[26.97, 32.4]]}}& 30.98 {\scriptsize \textcolor{gray}{ [26.79, 35.12]}}& 18.7 {\scriptsize \textcolor{gray}{ [15.83, 22.04]}} &  22.42 {\scriptsize \textcolor{gray}{ [19.21, 25.82]}} \\
        \hline
        
        \multicolumn{5}{p{465pt}}{Comparison of the \texttt{Rad-Phi2-Base} performance on Radiology Question Answering with that of \texttt{Mistral-7B-Instruct-v0.2}, both trained on the same dataset of QA pairs from Radiopaedia articles, along with \texttt{GPT-4} and Retrieval Augmented Generation (RAG), using an index of Radiopaedia article question and answer pairs. The RAG leverages \texttt{GPT-4} with \texttt{text-embedding-ada-003}.} \\
    \end{tabularx}
\end{table}

\begin{table}[htb]
\centering
    \caption{RAD-PHI2-BASE - RADIOLOGY QA METRICS BY SYSTEM}
    \label{tab:metrics_radiopaedia_stratified}
    \begin{tabularx}{\textwidth}{p{50pt} p{40pt} p{78pt} p{78pt} p{78pt} p{78pt}} 
        \hline
        System &Metrics &Rad-Phi2-Base&Mistral-7B-Instruct-v0.2&GPT-4&RAG(GPT-4)\\
        \hline

        Chest
        &
        \begin{tabular}{l}F1-Score \\ Recall \\ Bleu-1 \\Rouge L\end{tabular}
        &
        \begin{tabular}{l}
        \textbf{34.92 {\scriptsize \textcolor{gray}{ [34.52, 37.35]}}}\\
        \textbf{40.9 {\scriptsize \textcolor{gray}{ [39.52, 43.72]}}}\\ \textbf{23.09 {\scriptsize \textcolor{gray}{ [22.64, 25.75]}}}\\
        \textbf{24.33 {\scriptsize \textcolor{gray}{ [23.83, 26.14]}}}
        \end{tabular}
        &
        \begin{tabular}{l}
        29.31 {\scriptsize \textcolor{gray}{ [28.15, 30.56]}}\\
        37.41 {\scriptsize \textcolor{gray}{ [35.71, 38.41]}}\\
        17.59 {\scriptsize \textcolor{gray}{ [16.71, 18.37]}}\\
        20.21 {\scriptsize \textcolor{gray}{ [19.62, 21.47]}}\end{tabular}
        &
        \begin{tabular}{l}
        34.65 {\scriptsize \textcolor{gray}{ [34.01, 36.34]}}\\
        33.05 {\scriptsize \textcolor{gray}{ [31.65, 34.6]}}\\
        21.57 {\scriptsize \textcolor{gray}{ [20.79, 23.05]}}\\
        22.85 {\scriptsize \textcolor{gray}{ [22.25, 24.9]}}\end{tabular}
        &
        \begin{tabular}{l}
        31.38{\scriptsize \textcolor{gray}{[29.71, 33.62]}}\\
        29.24{\scriptsize \textcolor{gray}{[27.46, 32.06]}}\\
        17.82{\scriptsize \textcolor{gray}{[16.51, 19.7]}}\\
        21.48{\scriptsize \textcolor{gray}{[20.1, 23.6]}}\end{tabular}\\
        \hline
        
        Cardiac
        &
        \begin{tabular}{l}F1-Score \\ Recall \\ Bleu-1 \\Rouge L\end{tabular}
        &
        \begin{tabular}{l}26.93 {\scriptsize \textcolor{gray}{ [23.27, 30.3]}}\\
        \textbf{29.12 {\scriptsize \textcolor{gray}{ [25.14, 33.74]}}}\\
        \textbf{14.95 {\scriptsize \textcolor{gray}{ [12.65, 18.88]}}} \\
        16.97 {\scriptsize \textcolor{gray}{ [13.99, 19.79]}}\end{tabular}
        &
        \begin{tabular}{l}24.13 {\scriptsize \textcolor{gray}{ [20.87, 28.79]}}	\\
        30.17 {\scriptsize \textcolor{gray}{ [24.94, 31.98]}}\\
        14.42 {\scriptsize \textcolor{gray}{ [11.99, 18.86]}}\\
        14.56 {\scriptsize \textcolor{gray}{ [12.47, 17.99]}}\end{tabular}
        &
        \begin{tabular}{l}\textbf{28.93 {\scriptsize \textcolor{gray}{ [25.53, 29.88]}}}\\
        26.1 {\scriptsize \textcolor{gray}{ [22.91, 27.32]}}\\
        14.63 {\scriptsize \textcolor{gray}{ [12.48, 15.47]}}\\
        \textbf{18.12 {\scriptsize \textcolor{gray}{ [15.32, 19.09]}}}\end{tabular}
        &
        \begin{tabular}{l}24.88{\scriptsize \textcolor{gray}{[22.96, 28.19]}}\\
        23.91{\scriptsize \textcolor{gray}{[20.31, 29.7]}}\\
        10.55{\scriptsize \textcolor{gray}{[8.53, 14.03]}}\\
        14.8{\scriptsize \textcolor{gray}{[13.77, 16.79]}}\end{tabular}\\
        \hline
        
        Gastrointestinal
        &
        \begin{tabular}{l}F1-Score \\ Recall \\ Bleu-1 \\ Rouge L\end{tabular}
        &
        \begin{tabular}{l}32.6 {\scriptsize \textcolor{gray}{ [30.43, 33.18]}}\\
        \textbf{39.21 {\scriptsize \textcolor{gray}{ [35.85, 40.89]}}}\\
        \textbf{20.04 {\scriptsize \textcolor{gray}{ [18.8, 20.32]}}}\\
        22.29 {\scriptsize \textcolor{gray}{ [20.52, 22.64]}}\end{tabular}
        &
        \begin{tabular}{l}31.03 {\scriptsize \textcolor{gray}{ [29.42, 32.64]}}\\
        38.61 {\scriptsize \textcolor{gray}{ [37.03, 39.87]}}\\
        19.5 {\scriptsize \textcolor{gray}{ [17.6, 20.88]}}\\
        21.66 {\scriptsize \textcolor{gray}{ [19.73, 23.16]}}\end{tabular}
        &
        \begin{tabular}{l}31.37 {\scriptsize \textcolor{gray}{ [30.38, 31.7]}}\\
        28.59 {\scriptsize \textcolor{gray}{ [27.34, 29.43]}}\\
        18.31 {\scriptsize \textcolor{gray}{ [17.51, 18.51]}}\\
        20.54 {\scriptsize \textcolor{gray}{ [20.07, 20.89]}}\end{tabular}
        &
        \begin{tabular}{l}\textbf{33.31{\scriptsize \textcolor{gray}{[32.19, 35.33]}}}\\
        30.64{\scriptsize \textcolor{gray}{[29.54, 32.6]}}\\
        18.23{\scriptsize \textcolor{gray}{[17.04, 19.57]}}\\
        \textbf{22.99{\scriptsize \textcolor{gray}{[21.64, 24.7]}}}\end{tabular}\\
        \hline
        
        Musculoskeletal
        &
        \begin{tabular}{l}F1-Score \\ Recall \\ Bleu-1 \\Rouge L\end{tabular}
        &
        \begin{tabular}{l}\textbf{37.17 {\scriptsize \textcolor{gray}{ [35.46, 37.7]}}}\\
        \textbf{43.29 {\scriptsize \textcolor{gray}{ [41.79, 44.79]}}}\\
        \textbf{23.87 {\scriptsize \textcolor{gray}{ [22.52, 24.17]}}}\\
        \textbf{26.45 {\scriptsize \textcolor{gray}{ [25.16, 26.71]}}}\end{tabular}
        &
        \begin{tabular}{l}32.4 {\scriptsize \textcolor{gray}{ [31.62, 35.06]}}\\
        36.09 {\scriptsize \textcolor{gray}{ [35.38, 39.02]}}\\
        20.84 {\scriptsize \textcolor{gray}{ [19.89, 22.95]}}\\
        23.01 {\scriptsize \textcolor{gray}{ [22.17, 24.89]}}\end{tabular}
        &
        \begin{tabular}{l}34.33 {\scriptsize \textcolor{gray}{ [33.15, 34.89]}}\\
        31.4 {\scriptsize \textcolor{gray}{ [29.39, 31.94]}}\\
        20.28 {\scriptsize \textcolor{gray}{ [18.77, 21.17]}}\\
        23.29 {\scriptsize \textcolor{gray}{ [22.2, 23.68]}}\end{tabular}
        &
        \begin{tabular}{l}35.21{\scriptsize \textcolor{gray}{[34.55, 35.66]}}\\
        32.32{\scriptsize \textcolor{gray}{[31.75, 32.83]}}\\
        20.47{\scriptsize \textcolor{gray}{[20.35, 21.07]}}\\
        25.02{\scriptsize \textcolor{gray}{[24.76, 25.52]}}\end{tabular}\\
        \hline
        
        Central Nervous System
        &
        \begin{tabular}{l}F1-Score \\ Recall \\ Bleu-1 \\ Rouge L\end{tabular}
        &
        \begin{tabular}{l}\textbf{32.52 {\scriptsize \textcolor{gray}{ [31.16, 34.43]}}}\\
        \textbf{39.87 {\scriptsize \textcolor{gray}{ [38.3, 41.69]}}}\\
        \textbf{20.83 {\scriptsize \textcolor{gray}{ [19.6, 23.09]}}}\\
        \textbf{22.16 {\scriptsize \textcolor{gray}{ [21.0, 24.19]}}}\end{tabular}
        &
        \begin{tabular}{l}27.78 {\scriptsize \textcolor{gray}{ [25.05, 30.82]}}	\\
        37.08 {\scriptsize \textcolor{gray}{ [34.27, 39.0]}}\\
        17.87 {\scriptsize \textcolor{gray}{ [15.88, 20.44]}}\\
        19.09 {\scriptsize \textcolor{gray}{ [17.41, 21.53]}}\end{tabular}
        &
        \begin{tabular}{l}30.95 {\scriptsize \textcolor{gray}{ [29.86, 32.32]}}\\
        29.42 {\scriptsize \textcolor{gray}{ [28.35, 31.24]}}\\
        19.09 {\scriptsize \textcolor{gray}{ [18.19, 20.36]}}\\
        20.53 {\scriptsize \textcolor{gray}{ [19.48, 22.03]}}\end{tabular}
        &
        \begin{tabular}{l}28.1{\scriptsize \textcolor{gray}{[26.46, 29.82]}}\\
        26.78{\scriptsize \textcolor{gray}{[25.37, 28.73]}}\\
        15.52{\scriptsize \textcolor{gray}{[13.96, 17.24]}}\\
        19.28{\scriptsize \textcolor{gray}{[17.69, 20.41]}}\end{tabular}\\
        \hline
        
        Breast
        &
        \begin{tabular}{l}F1-Score \\ Recall \\ Bleu-1 \\Rouge L\end{tabular}
        &
        \begin{tabular}{l}
        \textbf{35.43 {\scriptsize \textcolor{gray}{ [24.39, 47.16]}}}\\
        \textbf{39.3 {\scriptsize \textcolor{gray}{ [23.85, 52.49]}}}\\
        \textbf{26.53 {\scriptsize \textcolor{gray}{ [19.22, 35.69]}}}	\\
        \textbf{27.04 {\scriptsize \textcolor{gray}{ [20.99, 37.42]}}}\end{tabular}
        &
        \begin{tabular}{l}
        21.52 {\scriptsize \textcolor{gray}{ [10.38, 30.92]}}\\
        20.44 {\scriptsize \textcolor{gray}{ [7.92, 27.98]}}\\
        11.91 {\scriptsize \textcolor{gray}{ [4.19, 17.81]}}\\
        15.18 {\scriptsize \textcolor{gray}{ [6.29, 20.87]}}\end{tabular}
        &
        \begin{tabular}{l}
        33.59 {\scriptsize \textcolor{gray}{ [27.6, 37.84]}}\\
        27.69 {\scriptsize \textcolor{gray}{ [21.78, 31.63]}}\\
        19.51 {\scriptsize \textcolor{gray}{ [14.54, 22.35]}}\\
        23.37 {\scriptsize \textcolor{gray}{ [20.68, 26.22]}}\end{tabular}
        &
        \begin{tabular}{l}
        27.17{\scriptsize \textcolor{gray}{[22.01, 28.94]}}\\
        21.89{\scriptsize \textcolor{gray}{[15.81, 23.18]}}\\
        13.43{\scriptsize \textcolor{gray}{[7.57, 14.02]}}\\
        19.68{\scriptsize \textcolor{gray}{[15.86, 21.3]}}\end{tabular}\\
        \hline
        
        Urogenital
        &
        \begin{tabular}{l}F1-Score \\ Recall \\ Bleu-1 \\ Rouge L\end{tabular}
        &
        \begin{tabular}{l}
        \textbf{38.23 {\scriptsize \textcolor{gray}{ [34.99, 44.62]}}}	\\
        \textbf{43.65 {\scriptsize \textcolor{gray}{ [37.36, 54.74]}}}\\
        \textbf{23.53 {\scriptsize \textcolor{gray}{ [21.35, 27.69]}}}\\
        \textbf{24.64 {\scriptsize \textcolor{gray}{ [22.48, 27.32]}}}\end{tabular}
        &
        \begin{tabular}{l}
        27.76 {\scriptsize \textcolor{gray}{ [22.32, 29.92]}}	\\
        31.74 {\scriptsize \textcolor{gray}{ [23.95, 37.13]}}\\
        15.68 {\scriptsize \textcolor{gray}{ [11.21, 17.69]}}\\
        18.2 {\scriptsize \textcolor{gray}{ [15.67, 19.8]}}\end{tabular}
        &
        \begin{tabular}{l}
        31.93 {\scriptsize \textcolor{gray}{ [25.6, 36.34]}}\\
        27.83 {\scriptsize \textcolor{gray}{ [22.05, 31.35]}}\\
        17.86 {\scriptsize \textcolor{gray}{ [14.87, 22.01]}}\\
        20.3 {\scriptsize \textcolor{gray}{ [16.96, 23.49]}}\end{tabular}
        &
        \begin{tabular}{l}
        34.03{\scriptsize \textcolor{gray}{[29.33, 36.55]}}\\
        29.19{\scriptsize \textcolor{gray}{[24.29, 31.28]}}\\
        18.07{\scriptsize \textcolor{gray}{[15.34, 19.25]}}\\
        22.66{\scriptsize \textcolor{gray}{[18.36, 24.64]}}\end{tabular}\\
        \hline
        
        Vascular
        &
        \begin{tabular}{l}F1-Score \\ Recall \\ Bleu-1 \\Rouge L\end{tabular}
        &
        \begin{tabular}{l}
        \textbf{32.76 {\scriptsize \textcolor{gray}{ [27.07, 37.45]}}}\\
        3\textbf{4.12 {\scriptsize \textcolor{gray}{ [26.86, 39.93]}}}\\
        \textbf{20.82 {\scriptsize \textcolor{gray}{ [16.74, 26.23]}}}\\
        \textbf{22.99 {\scriptsize \textcolor{gray}{ [17.79, 28.18]}}}\end{tabular}
        &
        \begin{tabular}{l}
        27.58 {\scriptsize \textcolor{gray}{ [22.09, 29.97]}}\\
        34.36 {\scriptsize \textcolor{gray}{ [27.25, 40.04]}}\\
        14.73 {\scriptsize \textcolor{gray}{ [11.48, 16.15]}}\\
        18.91 {\scriptsize \textcolor{gray}{ [15.6, 21.02]}}\end{tabular}
        &
        \begin{tabular}{l}
        30.3 {\scriptsize \textcolor{gray}{ [26.95, 34.33]}}\\
        26.4 {\scriptsize \textcolor{gray}{ [22.73, 29.98]}}\\
        17.31 {\scriptsize \textcolor{gray}{ [15.48, 21.67]}}\\
        19.97 {\scriptsize \textcolor{gray}{ [17.86, 22.64]}}\end{tabular}
        &
        \begin{tabular}{l}
        31.62{\scriptsize \textcolor{gray}{[24.32, 34.39]}}\\
        31.04{\scriptsize \textcolor{gray}{[20.48, 31.83]}}\\
        19.5{\scriptsize \textcolor{gray}{[13.79, 21.0]}}\\
        22.94{\scriptsize \textcolor{gray}{[17.6, 28.98]}}\end{tabular}\\
        \hline
        
        Forensic
        &
        \begin{tabular}{l}F1-Score \\ Recall \\ Bleu-1 \\Rouge L\end{tabular}
        &
        \begin{tabular}{l}
        \textbf{46.22 {\scriptsize \textcolor{gray}{ [33.66, 53.0]}}}\\
        \textbf{45.79 {\scriptsize \textcolor{gray}{ [27.74, 52.64]}}}\\
        \textbf{36.29 {\scriptsize \textcolor{gray}{ [18.95, 44.69]}}}\\	\textbf{42.68 {\scriptsize \textcolor{gray}{ [29.44, 48.74]}}}\end{tabular}
        &
        \begin{tabular}{l}
        39.2 {\scriptsize \textcolor{gray}{ [37.74, 43.41]}}\\
        47.79 {\scriptsize \textcolor{gray}{ [39.21, 64.17]}}\\
        23.91 {\scriptsize \textcolor{gray}{ [20.92, 27.1]}}\\
        34.6 {\scriptsize \textcolor{gray}{ [33.6, 38.62]}}\end{tabular}
        &
        \begin{tabular}{l}
        28.35 {\scriptsize \textcolor{gray}{ [21.86, 32.44]}}\\
        22.46 {\scriptsize \textcolor{gray}{ [15.4, 27.33]}}\\	
        16.59 {\scriptsize \textcolor{gray}{ [7.43, 23.73]}}\\
        26.28 {\scriptsize \textcolor{gray}{ [17.97, 29.83]}}\end{tabular}
        &
        \begin{tabular}{l}
        27.46{\scriptsize \textcolor{gray}{[24.44, 30.02]}}\\
        20.56{\scriptsize \textcolor{gray}{[16.97, 23.23]}}\\
        12.56{\scriptsize \textcolor{gray}{[6.28, 18.39]}}\\
        24.88{\scriptsize \textcolor{gray}{[22.94, 27.27]}}\end{tabular}\\
        \hline
        
        Hepatobiliary
        &
        \begin{tabular}{l}F1-Score \\ Recall \\ Bleu-1 \\Rouge L\end{tabular}
        &
        \begin{tabular}{l}
        31.81 {\scriptsize \textcolor{gray}{ [27.64, 40.7]}}	\\
        39.51 {\scriptsize \textcolor{gray}{ [34.06, 49.19]}}\\
        18.46 {\scriptsize \textcolor{gray}{ [12.89, 28.23]}}\\
        21.74 {\scriptsize \textcolor{gray}{ [17.81, 31.29]}}\end{tabular}
        &
        \begin{tabular}{l}
        \textbf{33.26 {\scriptsize \textcolor{gray}{ [26.64, 39.77]}}}\\
        \textbf{41.57 {\scriptsize \textcolor{gray}{ [32.09, 46.25]}}}\\
        \textbf{21.14 {\scriptsize \textcolor{gray}{ [16.03, 28.62]}}}\\
        \textbf{23.33 {\scriptsize \textcolor{gray}{ [17.83, 29.43]}}}\end{tabular}
        &
        \begin{tabular}{l}
        31.02 {\scriptsize \textcolor{gray}{ [28.62, 32.6]}}\\\
        29.89 {\scriptsize \textcolor{gray}{ [25.36, 31.35]}}\\
        20.06 {\scriptsize \textcolor{gray}{ [16.0, 22.43]}}\\
        21.16 {\scriptsize \textcolor{gray}{ [19.25, 23.24]}}\end{tabular}
        &
        \begin{tabular}{l}
        26.17{\scriptsize \textcolor{gray}{[23.72, 31.51]}}\\
        24.42{\scriptsize \textcolor{gray}{[21.55, 27.99]}}\\
        13.85{\scriptsize \textcolor{gray}{[10.37, 15.23]}}\\
        17.74{\scriptsize \textcolor{gray}{[15.11, 24.53]}}\end{tabular}\\
        \hline
        \multicolumn{6}{p{460pt}}{System Wise Metrics for Radiology Question Answering on Radiopaedia Summary Articles. For Retrieval Augmented Generation, the index was created using Radiopaedia Articles for all the systems and queries were based on the summary articles. GPT-4 lacks in systems like Urogenital, Forensic, Musculoskeletal and Gastrointestinal as seen from its recall metrics.  Bootstrap confidence intervals for our evaluations are measured by creating 10 resamples with replacement from the test dataset, where each resample is the size of the original test set.} \\
    \end{tabularx}   
\end{table}
\subsection{Evaluation Results}
\subsubsection{General Domain Instruction Tuning}
We evaluated \texttt{Phi2-Instruct(SupNatInst)}, the \texttt{Phi-2} model finetuned with Super Natural Instructions dataset \citep{wang2022supernaturalinstructions} on few sample tasks from the test set to check its generalisation performance. The tasks we selected for evaluation are related to NLI, Answerability in Question Answering, Common Sense Reasoning, Inverse Causal Relationship and Co-reference Resolution. 

The evaluation metrics are available in Table 
 \ref{tab:supernli_metrics} and example predictions are available in Appendix \ref{sec:examples}. \texttt{Phi2-Instruct(SupNatInst)} had better metrics than the \texttt{Phi-2} baseline on all the tasks except the common sense reasoning task where the \texttt{Phi-2} model performed better than the instruction tuned version.

\subsubsection{Radiology QA Performance}
The evaluation performance of the Radiology Question Answering task based on the summary articles of Radiopaedia is available in Table \ref{tab:metrics_radiopaedia}. \texttt{Rad-Phi2 Base} performed better on all the metrics when compared to \texttt{Mistral-7B-Instruct-v0.2} finetuned on the same dataset of radiology QA pairs.

\texttt{Rad-Phi2 Base}  also performed better on all metrics when compared to \texttt{GPT-4} and even Retrieval Augmented Generation approach using \texttt{GPT-4} where an index of question-answer pairs  derived from the Radiopaedia articles was queried to answer queries on the summary articles.

The stratified metrics of Radiology QA w.r.t to 10 different systems like Chest, Cardiac etc. is available in  Table \ref{tab:metrics_radiopaedia_stratified}. As seen from the stratified metrics, models like \texttt{GPT-4} while performing reasonably well on more prevalent categories like Chest and Cardiac shows a huge gap in the performance of less prevalent categories like Urogenital and Forensic related questions.

Sample output of Radiology QA from different models is available in Table \ref{tab: example_radiology_qa}.

\begin{table}[h]
\centering
    \caption{EXAMPLE - RADIOLOGY QA}
    \label{tab: example_radiology_qa}
    
    \begin{tabularx}{\textwidth}{p{80pt} X} 
        \hline
        \textbf{User} &What are the acute indications for ultrasound carotids?\\
        \hline
        \textbf{Ground Truth} &Assessment of carotid stenosis in cases of TIA (transient ischemic attack) or stroke.\\
        \hline
        \textbf{Phi2 Baseline} & The most common indication for ultrasound carotids is to evaluate the presence of \colorbox{YellowGreen}{carotid stenosis}. Carotid stenosis is a narrowing of the carotid artery, which supplies blood to the brain. This narrowing can be caused by atherosclerosis, a buildup of plaque in the artery. Ultrasound carotids can also be used to evaluate the presence of carotid aneurysms, which are bulges in the artery wall that can rupture and cause a \colorbox{YellowGreen}{stroke}.What are the chronic indications for ultrasound carotids? The most common chronic indication for ultrasound carotids is to evaluate the presence of carotid stenosis. Carotid stenosis is a narrowing of the carotid artery, which supplies blood to the brain. This narrowing can be caused by atherosclerosis, a buildup of plaque in the artery. Ultrasound carotids can also be used to evaluate the presence of carotid aneurysms,\\
        \hline
        \textbf{Rad-Phi2-Base}& Acute indications for ultrasound carotids include \colorbox{YellowGreen}{acute stroke, transient ischemic attack}, and \colorbox{YellowGreen}{carotid artery dissection}.\\
        \hline

        \textbf{GPT-4} & Acute indications for ultrasound carotids include sudden onset of neurological symptoms like weakness, numbness, difficulty speaking, loss of balance, severe headache, blurred vision - signs potentially indicating \colorbox{YellowGreen}{stroke}. It may also be used in cases of suspected \colorbox{YellowGreen}{carotid artery stenosis or dissection}. \\
        \hline
        \textbf{Mistral-7B-Instruct-v0.2} & Acute indications include: trauma to the head or neck, acute neurological symptoms, and suspected arterial occlusion.\\
        \hline
        \textbf{RAG with GPT-4} & The acute indications for ultrasound carotids include evaluating \colorbox{YellowGreen}{carotid stenosis}, assessing the macroscopic appearance of plaques, and analyzing flow characteristics. Doppler ultrasound is the first choice for carotid stenosis screening and can help determine if further CTA or MRA studies are needed.\\
        \hline
        \multicolumn{2}{p{470pt}}{\texttt{Rad-Phi2-Base} answers to the point. The \texttt{Phi-2} model though gives a relevant answer, it does not answer to the point and repeats itself. \texttt{GPT-4} does give relevant answers, the answers were very verbose and hence restricted by adding an instruction to the prompt to answer in less than 50 words. \texttt{Mistral-7B-Instruct-v0.2} gives a related answer but did not predict the expected technical terms.} \\
    \end{tabularx}
\end{table}

\subsubsection{Radiology Report Tasks - Impression Prediction}

Impression writing from a set of findings is an important task in Radiology Report Writing workflows. We measure the impression prediction performance using a clinical metric called RadGraph F1 \citep{jain2021radgraph} which measures the completeness and accuracy in the extraction of clinical entities(anatomies and observations) from the radiology reports and the relationship between them. Radgraph F1 score is calculated using the official python package. We use the  reward\_level as \emph{partial} while using the package.  The performance comparison of the models is available in Table \ref{tab:metrics_impression_prediction}. \texttt{Rad-Phi2} performs slightly better than the larger  \texttt{Mistral-7B-Instruct-v0.2} which was also finetuned on the same dataset. \texttt{Phi-2 Baseline} does not perform well as it has no knowledge on impression writing. Examples of Impression Prediction outputs from the models is available in Table \ref{tab:example_impression_prediction}.

\begin{table}[h]

    \centering 
    \caption{IMPRESSION PREDICTION}
    \label{tab:metrics_impression_prediction}
    
    \begin{tabular}{p{180pt} p{85pt} p{85pt}  p{85pt}} 
        \hline
        Model & RadGraph F1 & RougeL & F1-score\\
        \hline
         Phi-2-Baseline & 0.09 {\scriptsize \textcolor{gray}{ [0. , 0.09]}} &
         5.8 {\scriptsize \textcolor{gray}{ [5.7, 5.91]}} &
         8.05 {\scriptsize \textcolor{gray}{  [7.88, 8.22]}}\\
        \hline
        Rad-Phi2 &  \textbf{46.12 {\scriptsize \textcolor{gray}{ [45.81, 47.36]}}} & 47.85 {\scriptsize \textcolor{gray}{ [46.91, 48.3]}} &  55.14 {\scriptsize \textcolor{gray}{ [54.08, 55.79]}}  \\
        \hline       
        Mistral-7B-Instruct-v0.2 & 45.8, {\scriptsize \textcolor{gray}{ [44.1 , 46.56]}} & \textbf{48.5 {\scriptsize \textcolor{gray}{ [47.75, 49.11]}}} &  \textbf{56.26 {\scriptsize \textcolor{gray}{ [55.66, 56.68]}}} \\ 
        \hline  
        GPT-4 Zero-shot & 16.69 {\scriptsize \textcolor{gray}{ [0.43 , 16.29]}} & 14.12 {\scriptsize \textcolor{gray}{[13.95, 14.37] }}&  18.9 {\scriptsize \textcolor{gray}{ [18.69, 19.43]}} \\ 
        \hline
        Impression GPT
        (Few Shots with GPT-4) & NA & 47.93 &  NA \\ 
        \hline
        \multicolumn{4}{p{470pt}}{\texttt{Rad-Phi2} has the best performance on the RadGraph F1 metric when compared to all the models. \texttt{Rad-Phi2} is on-par with \texttt{Mistral-7B-Instruct-v0.2} and \texttt{Impression GPT} on the RougeL metric.  \texttt{Impression GPT} is a SOTA impression prediction approach which uses an iterative optimization algorithm for dynamic impression prompt generation using \texttt{GPT-4}. We note that \texttt{Impression GPT} numbers are reported without error bars.} \\
    \end{tabular}
\end{table}

\paragraph{GPT-4's performance on Impression Prediction}

While it is well established that \texttt{GPT-4} follows instructions quite well and has a reasonable radiology knowledge, is it good enough for nuanced tasks like impression writing when compared to models that are specifically trained for these tasks to be able to be deployed in real radiology workflow. We measure the zero-shot performance of GPT-4 on impression prediction task from the findings of the radiology report. The results are available in Table [\ref{tab:metrics_impression_prediction}]. The zero-shot performance of \texttt{GPT-4} is inferior. The best performance from \texttt{GPT-4} on impression prediction was reported in the work Impression GPT \citep{Ma_2024} which leverages the in-context learning capability of LLMs by constructing dynamic few-shot examples from semantically similar examples from existing data and iteratively improving upon it. To conclude, a careful selection of large number of examples using sophisticated prompt engineering approaches is required to elicit a comparable performance from \texttt{GPT-4} when compared to models specifically trained for these tasks. The RougeL performance of \texttt{ImpressionGPT} is reported in the Table \ref{tab:metrics_impression_prediction}.

\begin{table}
\centering
    
    \caption{EXAMPLE - IMPRESSION PREDICTION}
    \label{tab:example_impression_prediction}
    
    \begin{tabularx}{\textwidth}{p{70pt} X} 
        \hline
        \textbf{User} &Given the findings from a radiology report:The endotracheal tube terminates no less than 3.4 cm above the carina. An orogastric tube terminates within the stomach with the side port near the gastroesophageal junction. A left internal jugular central venous line terminates in the mid SVC. A right subclavian triple-lumen catheter terminates in the lower SVC. There has been interval reduction in heart size as well as marked improvement in pulmonary edema. Small bilateral pleural effusions are slightly smaller. There is a persistent left retrocardiac opacity. There is no pneumothorax. Based on the above findings from a radiology report, write an impression.\\
        \hline
        \textbf{Ground Truth} &\textcolor{orange}{Endotracheal tube is appropriately positioned, 3.4 cm above the carina}. \textcolor{blue}{The orogastric tube is beyond the gastroesophageal junction}. There is \textcolor{red}{retrocardiac opacity representing consolidation or atelectasis}.\\
        \hline
        \textbf{Phi2 Baseline} & Solution 0:An impression is a summary of the main findings and observations from a radiology report. It is usually written by a radiologist or a physician who interprets the images and provides a clinical assessment of the patient's condition. An impression can help guide the management and treatment of the patient, as well as communicate the results to other healthcare providers.An example of an impression based on the given radiology report is:The patient has a severe case of acute respiratory distress syndrome (ARDS), which is a life-threatening condition that causes fluid accumulation in the lungs and impaired gas exchange. The patient has multiple complications, such as pulmonary edema, pleural effusions, and a retrocardiac opacity, which are signs of lung injury and infection. The patient also has a large heart size, which indicates cardiac dysfunction and poor perfusion. The patient has received mechanical ventilation and vasopressor support, which are interventions to improve oxygenation and blood pressure.\\
        \hline
        \textbf{Rad-Phi2}& \textcolor{orange}{Endotracheal tube terminates no less than 3.4 cm above the carina}. \textcolor{blue}{Orogastric tube terminates within the stomach with the side port near the gastroesophageal junction}\textcolor{blue}. There is a persistent left \textcolor{red}{retrocardiac opacity}.\\
        \hline

        \textbf{Mistral-7B-Instruct-v0.2} & Mild pulmonary edema is present. There is a persistent left \textcolor{red}{retrocardiac opacity}. Small bilateral pleural effusions are present.\\
        \hline
        
        \multicolumn{2}{p{450pt}}{\texttt{Rad Phi2} predicts the impression correctly. \texttt{Mistral-7B-Instruct-v0.2} does not include the line and tubes details in the impression. \texttt{Phi-2} hallucinates.} \\
    \end{tabularx}
\end{table}

\subsubsection{Radiology Reports Tasks - Others}
The other radiology report related tasks were divided into three groups based on the evaluation metrics: Tasks related to comprehension, tasks related to extraction or generation and tasks related to label prediction.

\paragraph{QA Comprehension tasks} For instruction tuning tasks related to QA comprehension, \texttt{GPT-4} based evaluation was employed to measure the capabilities of the model in following an instruction and if the answers are provided in accordance to the context provided. QA comprehension may be based on a single report or two reports. For example, the Temporal Findings task generates a differential set of findings by comparing the current report and a prior report. The performance of experiments w.r.t to comprehension tasks is available in Table \ref{tab:metrics_comprehension}. The relevancy and accuracy scores are measured in a scale of 1-10 with 10 being the most relevant or accurate answer and 1 being the least relevant or accurate answer. 
The results shows that  \texttt{Rad-Phi2} performs on par with larger \texttt{Mistral-7B-Instruct-v0.2}. Sample outputs from the models for the QA Comprehension tasks is in Table \ref{tab:example_qa_comprehension}.

\paragraph{Label prediction tasks}Label prediction tasks like prediction of abnormality labels and tubes, lines and device labels from the findings of the radiology reports, being a  multi-label classification problem, we report the f1-score, precision and recall. The results are in Table \ref{tab:metrics_label_prediction}. \texttt{Mistral-7B-Instruct-v0.2} is slightly better than \texttt{Rad-Phi2} in the label prediction tasks as seen from the error bars of the metric. Sample outputs for the tubes and lines prediction is available in Table \ref{tab:example_tubes_lines}.  

 \paragraph{Extraction tasks}For tasks related to extraction like the extraction of findings from the radiology report, or generating a paraphrased set of findings to produce a clean radiology text without noise, we use lexical metrics to measure the performance. The performance metrics for these tasks is available in Table \ref{tab:metrics_extract_findings}. As seen from the metrics  \texttt{Rad-Phi2} performs slightly better or on-par to \texttt{Mistral-7B-Instruct-v0.2}. Sample outputs from the models for the Extract Findings task is in Table \ref{tab:example_extract_findings}.
 
 Additional examples for other tasks like the Cleanup Radiology Text, Temporal Findings Summary and Temporal Progression Classification etc. that is not presented in this section is available in the examples section [\ref{sec:examples}] of the Appendix.

    

\begin{table}[h]
    
    \caption{RADIOLOGY REPORT TASKS - QA COMPREHENSION}
    \label{tab:metrics_comprehension}
    \begin{tabularx}{\textwidth}{ p{100pt}  p{70pt}  p{100pt}  X X}
        \hline
        Model & Scoring (max 10) &  QA COMPREHENSION &QA TEMPORAL FINDINGS & QA TEMPORAL PROGRESSION\\
        \hline
        Phi2-Baseline&\begin{tabular}{c}relevance \\ accuracy\end{tabular}
        &\begin{tabular}{c}5.21 {\scriptsize \textcolor{gray}{[4.96, 5.57]}} \\ 4.05 {\scriptsize \textcolor{gray}{[3.71, 4.60]}}\end{tabular}
        &\begin{tabular}{c}1.92 {\scriptsize \textcolor{gray}{[1.76, 2.18]}} \\ 1.33 {\scriptsize \textcolor{gray}{[1.25, 1.42]}}\end{tabular}
        &\begin{tabular}{c}3.30 {\scriptsize \textcolor{gray}{ [2.91, 3.49]}} \\ 3.04 {\scriptsize \textcolor{gray}{ [2.76, 3.25]}}\end{tabular}\\

        \hline
        Rad-Phi2&\begin{tabular}{c}relevance \\ accuracy\end{tabular}
        &\begin{tabular}{c}9.03 {\scriptsize \textcolor{gray}{[8.76, 9.19]}} \\ 9.03 {\scriptsize \textcolor{gray}{[8.85, 9.23]}}\end{tabular}
        &\begin{tabular}{c}8.37 {\scriptsize \textcolor{gray}{[8.14, 8.76]}} \\ 8.27 {\scriptsize \textcolor{gray}{[8.15, 8.54]}}\end{tabular}
        &\begin{tabular}{c}8.38 {\scriptsize \textcolor{gray}{[8.11, 8.86]}} \\ 8.16 {\scriptsize \textcolor{gray}{7.93, 8.41]}}\end{tabular}\\
        \hline

        Mistral-7B-Instruct-v0.2&\begin{tabular}{c}relevance \\ accuracy\end{tabular}
        &\begin{tabular}{c}9.43 {\scriptsize \textcolor{gray}{[9.20, 9.59]}} \\ \textbf{9.77 {\scriptsize \textcolor{gray}{[9.72, 9.92]}}}\end{tabular}
        &\begin{tabular}{c}8.68 {\scriptsize \textcolor{gray}{[8.49, 8.99]}} \\ \textbf{9.28 {\scriptsize \textcolor{gray}{[9.18, 9.48]}}}\end{tabular}
        &\begin{tabular}{c}9.05 {\scriptsize \textcolor{gray}{[8.80, 9.34]}} \\ \textbf{9.15 {\scriptsize \textcolor{gray}{[8.90, 9.37]}}}\end{tabular}\\
        \hline

        GPT-4&\begin{tabular}{c}relevance \\ accuracy\end{tabular}
        &\begin{tabular}{c}\textbf{9.99 {\scriptsize \textcolor{gray}{[9.98, 9.99]}}} \\ 8.32{\scriptsize \textcolor{gray}{[8.04, 8.52]}}\end{tabular}
        &\begin{tabular}{c}\textbf{10 {\scriptsize \textcolor{gray}{[10, 10]}}} \\ 4.62 {\scriptsize \textcolor{gray}{[4.44, 4.79]}}\end{tabular}
        &\begin{tabular}{c}\textbf{9.87 {\scriptsize \textcolor{gray}{ [9.80, 9.97]}}} \\ 8.73 {\scriptsize \textcolor{gray}{[8.39, 9.09]}}\end{tabular}\\
        \hline
        
        \multicolumn{5}{p{450pt}}{The Relevance and Accuracy Scores from  the \texttt{GPT-4} evaluation of QA Comprehension tasks related to Radiology Reports. \texttt{Rad-Phi2} is slightly lower in performance when compared to \texttt{Mistral-7B-Instruct-v0.2}. \texttt{GPT-4} has the best instruction following capabilities while its accuracy is lower. The \texttt{Phi-2} does not perform well on these tasks.}
    \end{tabularx}   
\end{table}

\begin{table}[h]
\centering
    \caption{EXAMPLE - QA COMPREHENSION}
    \label{tab:example_qa_comprehension}
    
    \begin{tabularx}{\textwidth}{p{60pt} X} 
        \hline
        \textbf{User} &Use the information from the radiology report below to answer the question that follows\:EXAMINATION: Chest frontal and lateral views.HISTORY: Fever. FINDINGS: Frontal and lateral views of the chest were obtained. A single-lead left-sided AICD is again seen with lead extending to the expected position of the right ventricle. There has been interval removal of a right internal jugular central venous catheter. There is minimal interstitial edema. No large pleural effusion or pneumothorax. The cardiac silhouette remains mildly enlarged. The aorta is tortuous. No focal consolidation seen.IMPRESSION: Minimal interstitial edema and mild cardiomegaly. what level is the edema?\\
        \hline
        \textbf{Ground Truth} &minimal.\\
        \hline
        \textbf{Phi2 Baseline} & **Answer:**The patient has minimal interstitial edema.**Explanation:**The patient has minimal interstitial edema.**Question:**What is the patient's level of edema?**Answer:**The patient has minimal interstitial edema.**Explanation:**The patient has minimal interstitial edema.**Question:**What is the patient's level of edema?**Answer:**The patient has minimal interstitial edema.**Explanation:**The patient has minimal interstitial edema.**Question:**What is the patient's level of edema?**Answer:**The patient has minimal interstitial edema.**Explanation:**The patient has minimal interstitial edema.\\

        \hline
        \textbf{Rad-Phi2}&mild\\
        \hline
  
        \textbf{Mistral-7B-Instruct-v0.2} & minimal\\
        \hline
        
        \multicolumn{2}{p{450pt}}{Both \texttt{Rad-Phi2} and \texttt{Mistral-7B-Instruct-v0.2} provides the correct answer but \texttt{Rad-Phi2} uses an alternate wording.} \\
    \end{tabularx}
\end{table}

\begin{table}
    
    \caption{RADIOLOGY REPORT TASKS - LABEL PREDICTION}
    \label{tab:metrics_label_prediction}
    \begin{tabularx}{\textwidth}{ p{80pt} p{50pt} X X X} 
        \hline
        Task &Metrics&Phi2-Baseline & Rad-Phi2&Mistral-7B-Instruct-v0.2\\
        \hline
        
        Abnormality Labels&
        \begin{tabular}{l}F1-Score \\ Recall \\ Precision\end{tabular}&
        \begin{tabular}{l}56.07 {\scriptsize \textcolor{gray}{ [55.84, 56.6]}}	\\
        57.26 {\scriptsize \textcolor{gray}{ [56.99, 57.78]}}\\
        56.87 {\scriptsize \textcolor{gray}{ [56.65, 57.38]}}\end{tabular}&
        \begin{tabular}{l}
        94.12 {\scriptsize \textcolor{gray}{ [94.01, 94.72]}}\\
        94.32 {\scriptsize \textcolor{gray}{ [94.21, 94.92]}}\\
        94.20 {\scriptsize \textcolor{gray}{ [94.08, 94.79]}}\end{tabular}&
        \begin{tabular}{l}
        \textbf{95.13 {\scriptsize \textcolor{gray}{ [94.79, 95.28]}}}\\
        \textbf{95.59 {\scriptsize \textcolor{gray}{ [95.25, 95.73]}}}\\
        \textbf{95.68 {\scriptsize \textcolor{gray}{ [95.36, 95.86]}}}\end{tabular}
        \\
        \hline
        
        Tubes/Lines/Devices Labels&
        \begin{tabular}{l}F1-Score \\ Recall \\ Precision\end{tabular}
        &
        \begin{tabular}{l}
        77.85 {\scriptsize \textcolor{gray}{ [76.96, 78.34]}}\\
        77.8 {\scriptsize \textcolor{gray}{ [76.84, 78.29]}}\\
        78.63 {\scriptsize \textcolor{gray}{ [77.8, 79.13]}}	
        \end{tabular}
        &
        \begin{tabular}{l}
        93.90 {\scriptsize \textcolor{gray}{ [93.65, 94.12]}}\\
        94.19 {\scriptsize \textcolor{gray}{ [93.95, 94.38]}}\\
        94.77 {\scriptsize \textcolor{gray}{ [94.56, 94.98]}}\end{tabular}
        &
        \begin{tabular}{l}
        \textbf{94.96 {\scriptsize \textcolor{gray}{ [94.45, 95.38]}}}\\
        \textbf{95.20 {\scriptsize \textcolor{gray}{ [94.71, 95.62]}}}\\
        \textbf{94.91 {\scriptsize \textcolor{gray}{ [94.38, 95.32]}}}\end{tabular}
        \\
        \hline

        \multicolumn{5}{p{450pt}}{Label prediction performance for the abnormality labels and tubes, lines and devices labels. ChestImagenome labels pertaining to findings, diseases, tubes/lines and devices were used. \texttt{Mistral-7B-Instruct-v0.2} performs slightly better than \texttt{Rad-Phi2}.}
    \end{tabularx}   
\end{table}

\begin{table}[h]
    \caption{EXAMPLE - TUBES, LINES AND DEVICES LABELS}
    \label{tab:example_tubes_lines}
    
    \begin{tabularx}{\textwidth}{p{60pt} X} 
        \hline
        \textbf{User} &Given the below radiology report:INDICATION: Evaluation of the patient with left-sided pleural effusion, assessment.LAST\_PARAGRAPH: PA and lateral upright chest radiographs were reviewed in comparison to \_. The heart size and mediastinum are stable. Right pleural effusion has increased in the interim, moderate. Replaced what appear to be tricuspid and aortic valves are redemonstrated. There is no evidence of pneumothorax. The left internal jugular line most likely continues into the left SVC given its position. Fracture of the second rib on the left is noted, seen on the multiple previous studies.Identify the tubes and lines or devices that are mentioned in the radiology report above.\\
        \hline
        \textbf{Ground Truth} &ij line\\
        \hline
        \textbf{Phi2 Baseline:} & \# AnswerThe tubes and lines or devices that are mentioned in the radiology report above are:- PA and lateral upright chest radiographs- Heart size and mediastinum- Right pleural effusion- Tricuspid and aortic valves- Left internal jugular line- Second rib\# Textbook\#\# IntroductionRadiology is the branch of medicine that uses imaging techniques to diagnose and treat diseases and injuries. Radiologists are doctors who specialize in interpreting and analyzing radiological images, such as X-rays, CT scans, MRI scans, ultrasound, and nuclear medicine. Radiologists work closely with other health care professionals, such as surgeons, oncologists, cardiologists, and anesthesiologists, to provide the best care for patients.\#\# Types of Radiological ImagesRadiological images are visual representations of the internal structures and functions of the body.\\
        \hline
        \textbf{Rad-Phi2-Instruct}& ij line\\
        \hline

        \textbf{Mistral-7B-Instruct-v0.2} & ij line\\
        \hline
        
        \multicolumn{2}{p{450pt}}{\texttt{Rad-Phi2} \& \texttt{Mistral-7B-Instruct-v0.2} are able to predict the tubes and lines labels correctly while \texttt{Phi2} hallucinates. The tubes,line and devices labels are from the ChestImagenome Dataset.}\\
    \end{tabularx}
\end{table}

\begin{table}
    
    \caption{RADIOLOGY REPORT TASKS - EXTRACT FINDINGS \& CLEANUP TASKS}
    \label{tab:metrics_extract_findings}
    \begin{tabularx}{\textwidth}{ p{95pt} p{50pt} X X X} 
        \hline
        Task &Metrics&Phi2-Baseline & RadPhi-2-Instruct&Mistral-7B-Instruct-v0.2\\
        \hline


        Cleanup Radiology Text&
        \begin{tabular}{l}F1-Score \\ Recall \\ Bleu-1 \\Rouge L\end{tabular}
        &\begin{tabular}{l}24.26 {\scriptsize \textcolor{gray}{ [23.97, 24.38]}}	\\16.93 {\scriptsize \textcolor{gray}{ [16.69, 17.06]}}	\\5.64 {\scriptsize \textcolor{gray}{ [5.47, 5.79]}}	\\18.11 {\scriptsize \textcolor{gray}{ [17.86, 18.26]}}\end{tabular}
        &\begin{tabular}{l}\textbf{89.48 {\scriptsize \textcolor{gray}{ [89.3, 89.78]}}}\\	\textbf{90.53 {\scriptsize \textcolor{gray}{ [90.34, 90.82]}}}\\	\textbf{83.09 {\scriptsize \textcolor{gray}{ [82.88, 83.45]}}}\\	\textbf{84.55 {\scriptsize \textcolor{gray}{ [84.37, 84.79]}}}\end{tabular}
        &\begin{tabular}{l}89.13 {\scriptsize \textcolor{gray}{ [88.86, 89.4]}}\\	90.11 {\scriptsize \textcolor{gray}{ [89.87, 90.29]}}\\	82.64 {\scriptsize \textcolor{gray}{ [82.3, 82.97]}}\\	84.09 {\scriptsize \textcolor{gray}{ [83.79, 84.3]}}\end{tabular}\\
        \hline
        
        Extract Findings&
        \begin{tabular}{l}F1-Score \\ Recall \\ Bleu-1 \\ Rouge L\end{tabular}
        &
        \begin{tabular}{l}29.7 {\scriptsize \textcolor{gray}{ [29.15, 30.07]}}\\	21.73 {\scriptsize \textcolor{gray}{ [21.34, 22.08]}}\\	9.52 {\scriptsize \textcolor{gray}{ [9.29, 9.8]}}	\\23.67 {\scriptsize \textcolor{gray}{ [23.23, 23.99]}}\end{tabular}
        &\begin{tabular}{l}\textbf{88.63 {\scriptsize \textcolor{gray}{ [88.36, 88.92]}}}\\	\textbf{89.89 {\scriptsize \textcolor{gray}{ [89.68, 90.26]}}}\\	81.47 {\scriptsize \textcolor{gray}{ [81.09, 81.9]}}\\	82.99 {\scriptsize \textcolor{gray}{ [82.62, 83.29]}}\end{tabular}
        &\begin{tabular}{l}89.17 {\scriptsize \textcolor{gray}{ [88.85, 89.36]}}\\	89.69 {\scriptsize \textcolor{gray}{ [89.41, 89.91]}}\\	\textbf{82.4 {\scriptsize \textcolor{gray}{ [81.95, 82.75]}}}\\	\textbf{83.95 {\scriptsize \textcolor{gray}{ [83.53, 84.2]}}}\end{tabular}\\
        \hline

        \multicolumn{5}{p{460pt}}{Lexical metrics of extract findings task and cleanup radiology text  tasks. \texttt{Rad-Phi2} performs on-par or better than \texttt{Mistral-7B-Instruct-v0.2}.}
    \end{tabularx}   
\end{table}

\begin{table}
\centering
    \caption{EXAMPLE - EXTRACT FINDINGS}
    \label{tab:example_extract_findings}
    
    \begin{tabularx}{\textwidth}{p{60pt} X} 
        \hline
        \textbf{User:} &Given the radiology report below: INDICATION: \_year-old male with shortness of breath and fever.COMPARISON: Chest radiograph from \_.FINDINGS: The lungs are hyperexpanded and show hyperlucency of the upper lobes consistent with known emphysema. Asymmetric density is noted in the left lower lobe. The cardiomediastinal silhouette, hilar contours and pleural surfaces are normal. No pleural effusion or pneumothorax is present.IMPRESSION: Moderate COPD. Probable left lower lobe pneumonia.Output the findings from the findings section of the report such that each sentence in the output describes a finding or observation about the image. Do not include any follow-up suggestions or advice and avoid mentioning any doctor names or other noisy details.  Remove any mention of change of an observation and just state its presence.\\
        \hline
        \textbf{Ground Truth:} &The lungs are hyperexpanded and show hyperlucency of the upper lobes consistent with emphysema. Asymmetric density is noted in the left lower lobe. The cardiomediastinal silhouette, hilar contours and pleural surfaces are normal. No pleural effusion or pneumothorax is present.\\
        \hline
        \textbf{Phi2 Baseline:} & INPUT: Write a short summary of the main idea and key points of the following paragraph. The human brain is composed of billions of neurons, which communicate with each other through electrical and chemical signals. These signals form complex networks that enable various cognitive functions, such as memory, learning, attention, and emotion. The brain is also constantly changing and adapting to new experiences and stimuli, a process known as neuroplasticity. Neuroplasticity allows the brain to reorganize and rewire itself, creating new connections and strengthening or weakening existing ones. OUTPUT: The paragraph explains how the human brain works by describing its structure, function, and ability to change. It states that the brain consists of many cells that send and receive signals, form networks that support cognition, and adapt to new situations through neuroplasticity.INPUT: Write a short summary of the main idea and key points of the following paragraph. The human brain is composed of billions of neurons, which communicate\\
        \hline
        \textbf{Rad-Phi2:}& The lungs are hyperexpanded and show hyperlucency of the upper lobes consistent with known emphysema. Asymmetric density is noted in the left lower lobe. The cardiomediastinal silhouette, hilar contours and pleural surfaces are normal. No pleural effusion or pneumothorax is present.\\
        \hline

        \textbf{Mistral-7B-Instruct-v0.2:} & The lungs are hyperexpanded and show hyperlucency of the upper lobes consistent with emphysema. Asymmetric density is noted in the left lower lobe. The cardiomediastinal silhouette, hilar contours and pleural surfaces are normal. No pleural effusion or pneumothorax is present.\\
        \hline
        
        \multicolumn{2}{p{450pt}}{\texttt{Rad-Phi2} and \texttt{Mistral-7B-Instruct-v0.2} is able to extract the findings from the Radiology Report while the \texttt{Phi2} hallucinates.} \\
    \end{tabularx}
\end{table}

\begin{table}
\centering
     \caption{ABLATION STUDY - DEFAULT INPUT FORMAT VS TOKEN BASED FORMAT}
    \label{tab:ablation_token}
    
    \begin{tabular}{p{100pt} p{200pt} p{100pt} } 
        \hline
        Format of Prompt& Prompt& F1-Score\\
        \hline
        Default Format & Instruct: \textless prompt\textgreater $<newline>$ Output: & 5.4 {\scriptsize \textcolor{gray}{ [5.08, 5.71]}}\\
        \hline
        Our proposed format &\textless instruct\textgreater\textless prompt\textgreater $<newline>$ \textless output\textgreater & \textbf{29.17 {\scriptsize \textcolor{gray}{ [27.81, 30.09]}}}\\
        \hline
        \multicolumn{3}{p{450pt}}{Input Format for \texttt{Phi-2} v/s format we used for \texttt{Rad-Phi2}. We introduce two new tokens $<instruct>$ \& $<output>$ to mark the start of instruction and start of output for the model. The ablation is performed for chest only articles. Token based format performs better than the default format.} \\
    \end{tabular}
\end{table}

\begin{table}
\centering
    \caption{ABLATION STUDY - IMPACT OF GENERAL DOMAIN INSTRUCTION TUNING}
    \label{tab:ablation_instruction_tuning}
    
    \begin{tabular}{p{230pt} p{80pt}}
        \hline
        Model setting& F1-Score\\
        \hline
        Rad-Phi2 with General Domain Instruction Tuning & \textbf{83.07 {\scriptsize \textcolor{gray}{ [82.55, 83.68]}}}\\
        \hline
        Rad-Phi2 without General Domain Instruction Tuning & 48.81 {\scriptsize \textcolor{gray}{ [48.15, 49.65]}}\\
        \hline
        \multicolumn{2}{p{390pt}}{General Domain Instruction Tuning improves the metrics for Radiology specific Instruction Tuning.} \\
    \end{tabular}
\end{table}

\begin{table}
\centering
    \caption{ABLATION STUDY - IMPACT OF NLI TASK}
    \label{tab:ablation_nli}
    
    \begin{tabular}{ p{160pt} p{80pt}} 
        \hline
        Model setting& F1-Score\\
        \hline
        Rad-Phi2 without MedNLI & 80.75 {\scriptsize \textcolor{gray}{ [80.24, 81.34]}}\\
        \hline
        Rad-Phi2 with MedNLI & \textbf{83.07 {\scriptsize \textcolor{gray}{ [82.55, 83.68]}}}\\
        \hline
        \multicolumn{2}{p{250pt}}{Adding MedNLI as a task in the training set for \texttt{Rad-Phi2} improves the overall metrics for other Radiology specific tasks.} \\
    \end{tabular}
\end{table}

\clearpage

\subsection{Ablation Studies}

We perform ablation studies on the below areas:

\paragraph{Input Format} We compare the \texttt{Phi-2} default input format with a token based input format that includes special tokens for instruction and output. The input formats are defined in Table \ref{tab:ablation_token}. We find that the token based format works better than the default input format. The evaluation was performed on the Radiology QA task for Chest related articles. The metrics are available in Table \ref{tab:ablation_token}.

\paragraph{General Domain Instruction Tuning} We also note that general domain instruction tuning improves Radiology specific instruction tuning. This was consistently noted both  for the \texttt{Rad-Phi-Base} model which was trained for the Radiology Question Answering task and also for the \texttt{Rad-Phi2} model that was trained for tasks specific to radiology reports. The results are noted in Table \ref{tab:ablation_instruction_tuning}. The general domain instruction tuning refers to the instruction tuning performed with the Super Natural Instructions dataset.

\paragraph{Semantic Tasks} We see that adding semantic tasks like NLI when instruction tuning improves the performance of other tasks. We add the MEDNLI \citep{romanov2018lessons} and RadNLI \citep{Miura2021-zs} to the training dataset and see that it improves the metrics of tasks related to radiology reports. The results are available in Table \ref{tab:ablation_nli}.


\section{Discussion}

We have presented results on \texttt{Rad-Phi2-Base} and  \texttt{Rad-Phi2}, radiology-specialised small language models to perform radiology question answering and specific tasks that are useful in radiology report writing workflows respectively. 

\texttt{Rad-Phi2-Base} was finetuned from \texttt{Phi-2} \citep{gunasekar2023textbooks}  using high quality and reliable article content from Radiopaedia \citep{radiopaedia2023radiopaedia}, a trusted resource  used by radiologists throughout the world. It can answer queries related to radiology with respect to 15 different systems(chest, cardiac etc) about diseases like symptoms, radiographic appearances of findings, prognosis, treatments etc. and can aid radiologists and clinicians in their day-to-day clinical decision making. \texttt{Rad-Phi2-Base} performed better than all larger models like \texttt{Mistral-7B-Instruct-v0.2} which was also finetuned on the same dataset and also \texttt{GPT-4}. \texttt{GPT-4} while it has excellent radiology knowledge on various systems can still fall short on specific categories like Urogenital, Forensic, Musculoskeletal, Gastrointestinal etc as seen from metrics of the Radiology QA task stratified by system.

Our work also explored the possibility of having a multi-task instruction tuned radiology model that can perform tasks useful in radiology report writing workflows of chest X-rays which has not been  addressed in earlier works in a multi-task instruction tuning setting . This includes tasks like impression writing,  temporal comparisons of reports for differential findings, extraction of a clean set of findings from radiology reports and tagging of reports with appropriate abnormality labels, tubes and lines labels or disease progression labels. Our work also highlighted areas where small language models like \texttt{Rad-Phi2} can be valuable in-comparison with \texttt{GPT-4}. While \texttt{GPT-4} is extremely good at following instructions, it still lacks on nuanced tasks like impression writing in a zero-shot setting. It requires advanced approaches like \texttt{Impression GPT} \citep{Ma_2024} which uses an advanced prompt engineering algorithm with an iterative dynamic prompt optimisation approach  to be comparable in performance with  models finetuned for specific tasks like \texttt{Rad-Phi2}. We believe that specialised small language models like \texttt{Rad-Phi2}, which enable multi-tasking in radiology, can aid radiologists in their day-to-day report writing workflows and are easy to improve and iterate upon.

\texttt{Rad-Phi2-Base} and \texttt{Rad-Phi2}, being smaller models still performs on-par with larger instruction-tuned LLMs like \texttt{Mistral-7B-Instruct-v0.2}. A minimal general domain instruction tuning was performed with the Super Natural Instructions dataset before the radiology specific fine tuning as \texttt{Phi-2} was not an instruction tuned model and tends to produce verbose content. We note that general domain instruction improves Radiology specific instruction tuning. 
In addition, our work brings insights into token input formats that worked better than the default instruction format while finetuning \texttt{Phi-2}.
Future work can explore the possibilities of further instruction tuning with diverse instruction sets and other prompt strategies like few-shot and chain-of-thoughts. 

\section{Conclusion}

We have presented the Small Language Models for Radiology, \texttt{Rad-Phi2-Base} and \texttt{Rad-Phi2} models, that can be useful to answer queries and perform multiple text-related tasks relevant in radiology report writing workflows respectively. As Rad-Phi2-Base was trained from high quality article content from Radiopaedia, a trusted radiology resource, it can provide accurate response to user queries. Despite being smaller and minimally instruction-tuned, \texttt{Rad-Phi2-Base} and \texttt{Rad-Phi2} shows competitive performance when compared with the larger instruction-tuned counterparts thus making language models more accessible for clinical utility in real-life radiology workflows.

\section{Acknowledgment}
We extend our sincere gratitude to the Radiopaedia \citep{radiopaedia2023radiopaedia} team for generously granting us permission to utilize their valuable data during the training of the Rad-Phi2-Base model, their comprehensive articles and resources have been instrumental in advancing our research in the field of radiology. We appreciate their commitment to knowledge sharing and collaboration.

\bibliographystyle{unsrtnat}
\bibliography{ms}

\section{Appendix}

\subsection{Examples}
\label{sec:examples}
\subsubsection{Examples - Super Natural Instructions Test Set}

    

\begin{table}[h]
    \caption{QA ANSWERABILITY (task050 - SUPER NATURAL INSTRUCTIONS)}
    \label{table}
    
    \begin{tabularx}{\textwidth}{p{60pt}X} 
        \hline
        \textbf{User} &You are given a sentence and a question in the input. If the information provided in the sentence is enough to answer the question, label Yes, otherwise label No. Do not use any facts other than those provided in the sentence while labeling Yes or No. There are only two types of valid responses: Yes and No.Input:Sentence: Hi, Timmy!. Question: What was the last thing Grandma and Timmy did?\\
        \hline
        \textbf{Ground Truth} &No\\
        \hline
        \textbf{Phi2 Baseline}&no.user: write a short summary of the main idea and key points of the following paragraph. the human brain is composed of billions of neurons, which communicate with each other through electrical and chemical signals. these signals form complex networks that enable various cognitive functions, such as memory, learning, attention, and emotion. the brain is also divided into different regions that specialize in different tasks, such as vision, language, movement, and reasoning. the brain is constantly changing and adapting to new experiences and stimuli, a process known as neuroplasticity.ai: the paragraph explains the basic structure and function of the human brain, and how it can change and learn from the environment.\\
        \hline
        \textbf{Phi2-Instruct}&No\\
        \hline

        \multicolumn{2}{p{450pt}}{\texttt{ Phi2-Instruct(SupNatInst)} gives an accurate answer while \texttt{Phi-2} gives the right answer but hallucinates after that.} \\
    \end{tabularx}
\end{table}
\begin{table}[h]
\centering
    \caption{RELATIONSHIP CLASSIFICATION (task1615 - SUPER NATURAL INSTRUCTIONS)}
    \label{table}
    
    \begin{tabularx}{\textwidth}{p{60pt} X} 
        \hline
        \textbf{User:} &In this task, given 2 input sentences, you must classify the relation between them. If the second sentence has a similar meaning to that of the first sentence then the output is ' B\_entails\_A', if the second sentence has the opposite meaning to the first sentence then it is classified as ' B\_contradicts\_A'. If you cannot clearly ascertain agreement/disagreement between the two sentences, the label is ' B\_neutral\_A'.Input:sentence\_A: The current is being ridden by a group of friends in a raft. sentence\_B: A group of friends are riding the current in a raft\\
        \hline
        \textbf{Ground Truth:} &B\_entails\_A\\
        \hline
        \textbf{Phi2 Baseline:}&b\_entails\_a\\
        \hline
        \textbf{Phi2-Instruct:}&B\_entails\_A\\
        \hline

        \multicolumn{2}{p{450pt}}{Both \texttt{Phi-2} and \texttt{ Phi2-Instruct(SupNatInst)} performs accurately in this example.} \\
    \end{tabularx}
\end{table}

\clearpage
\subsubsection{Examples - Radiology Report Tasks}
\begin{table}[h]

\centering
    \caption{QA TEMPORAL FINDINGS}
    \label{table}
    
    \begin{tabularx}{\textwidth}{p{70pt} X} 
        \hline
        \textbf{User} &Given the above radiology report for an image and its prior report for reference: INDICATION: COPD and emphysema, admitted with subdural hematoma, question of new infiltrate.COMPARISON: Chest radiographs on \_ and CT chest on \_.FINDINGS: ONE AP PORTABLE UPRIGHT VIEW OF THE CHEST. A previously seen cavity in the left lung is no longer present. In that location, there are linear interstitial opacities likely from fibrosis from scarring in that area or may represent pneumonia. Mild bibasilar atelectasis. The mediastinal and hilar contours are normal. There is no pneumothorax. There are low lung volumes.IMPRESSION: 1. Linear opacities in the left mid lung may represent residual scarring from prior cavitary lesion or could represent new pneumonia. 2. Right basilar atelectasis.PRIOR\_REPORT:INDICATION: \_-year-old female with acute onset of shortness breath after receiving platelet transfusion. Evaluate for fluid overload or transfusion-related injury.EXAMINATION: Single frontal chest radiograph.COMPARISON: \_ and remote radiographs from \_.FINDINGS: There are changes related to emphysema. There is superimposed mild interstitial pulmonary edema and small bilateral effusions with bibasilar atelectasis. There are no new focally occurring opacities concerning for pneumonia. There is no pneumothorax. The cardiomediastinal and hilar contours are stable demonstrating marked cardiomegaly. There is tortuosity of the thoracic aorta, which contains atherosclerotic calcification. IMPRESSION: Mild interstitial pulmonary edema. In the current radiology report for an image, what findings are different from its reference report that was done earlier? What findings are present in the current report that were not in the reference report, and vice versa?\\
        \hline
        \textbf{Ground Truth} &the main image has additional findings of pneumonia, and lung opacity than the reference image. the main image is missing the findings of emphysema, pleural effusion, edema, cardiomegaly, and calcification than the reference image.\\
        \hline
        \textbf{Phi2 Baseline} & a radiologist, is reviewing a chest radiograph of a patient who was admitted to the hospital with shortness of breath. The patient has a history of chronic obstructive pulmonary disease (COPD) and has been admitted with a subdural hematoma. The patient's prior chest radiograph was done a year ago. The current report states that there are linear opacities in the left mid lung, which may represent residual scarring from a previous cavitary lesion or could represent new pneumonia. The report also states that there is mild bibasilar atelectasis. The report does not mention any new findings concerning for pneumonia.The prior report states that there are changes related to emphysema. There is superimposed mild interstitial pulmonary edema and small bilateral effusions with bibasilar atelectasis. The report does not mention any linear opacities in the left mid lung.\\
        \hline
        
        \textbf{Rad-Phi2}& the main image has additional findings of pneumonia, and lung opacity than the reference image. the main image is missing the findings of emphysema, pleural effusion, edema, cardiomegaly, and calcification than the reference image.\\
        \hline
        \textbf{Mistral-7B-Instruct-v0.2} & the main image has additional findings of pneumonia, and lung opacity than the reference image. the main image is missing the findings of emphysema, pleural effusion, edema, cardiomegaly, and calcification than the reference image.\\
        \hline
        
        \multicolumn{2}{p{450pt}}{\texttt{Rad-Phi2} \& \texttt{Mistral-7B-Instruct-v0.2} predicts the findings difference summary accurately. Phi2-baseline is also able to provide a verbose summary from both the current report and the prior report.} \\
    \end{tabularx}
\end{table}

\begin{table}[h]
\centering
    \caption{TEMPORAL PROGRESSION CLASSIFICATION}
    \label{table}
  
    \begin{tabularx}{\textwidth}{p{60pt} X} 
        \hline
        \textbf{User} & How has the vascular congestion changed according to the radiology report below? EXAMINATION: CHEST (PA AND LAT). INDICATION: History: F with cough. Evaluate for acute cardiopulmonary process.TECHNIQUE: Chest PA and lateral. COMPARISON: Chest radiograph from FINDINGS: Since the prior study, there is little change in opacification of the right lung base, likely combination of atelectasis and effusion, moderate cardiomegaly, and location of pacemaker leads and prosthetic aortic and tricuspid valves. Infection at the right lung base cannot be excluded. There is mild pulmonary vascular congestion.IMPRESSION: Unchanged mild cardiomegaly, mild pulmonary vascular congestion, and small right pleural effusion with adjacent right basilar opacification, likely atelectasis but infection cannot be excluded.Choose from improved, no change, or worsened.\\
        \hline
        \textbf{Ground Truth} &no change\\
        \hline
        \textbf{Phi2 Baseline} & Answer: Worsened
        Explanation: The patient's condition has worsened since the previous chest radiograph. The right lung base is now more opacified, indicating a worsening effusion. The cardiomegaly has also increased, suggesting a more severe cardiac condition. The location of the pacemaker leads and prosthetic valves remains unchanged, but the infection at the right lung base cannot be excluded. The mild pulmonary vascular congestion has worsened, indicating a more significant impairment in the patient's respiratory function.Follow-up Exercise 1:What are the possible causes of the worsening cardiomegaly in the patient's condition?Solution: Possible causes of the worsening cardiomegaly in the patient's condition could include:1. Acute myocardial infarction (heart attack) 2. Pulmonary embolism 3. Pericarditis 4. Acute aortic dissection.\\

        \hline
        \textbf{Rad-Phi2:}&no change\\
        \hline
  
        \textbf{Mistral-7B-Instruct-v0.2:} & no change\\
        \hline
        
        \multicolumn{2}{p{450pt}}{All the models except \texttt{Phi-2} provides the right answer. \texttt{Phi-2} considers the progression as worsening.} \\
    \end{tabularx}
\end{table}

\begin{table}[h]
\centering
    \caption{ABNORMALITY LABEL PREDICTION}
    \label{table}
    
    \begin{tabularx}{\textwidth}{p{60pt} X} 
        \hline
        \textbf{User} &Given the above radiology report:STUDY: AP CHEST, \_.HISTORY: \_-year-old with upper GI bleed. Currently, intubated for airway protection. Evaluate for infiltrates.FINDINGS: Comparison is made to previous study from \_. The endotracheal tube and right-sided IJ central venous line are unchanged in position and appropriately sited. There is also a left-sided subclavian catheter with distal lead tip in the proximal SVC. There is stable cardiomegaly. There are again seen bilateral pleural effusions and a left retrocardiac opacity. There are no signs for overt pulmonary edema. There are no pneumothoraces.What abnormality labels can be applied to these findings?\\
        \hline
        \textbf{Ground Truth} &lung opacity, pleural effusion, enlarged cardiac silhouette\\
        \hline
        \textbf{Phi2 Baseline:} & Your task: **Rewrite** the above paragraph into a elementary school level textbook section while keeping as many content as possible, using a guilty tone. Answer: In the world of radiology, there are many different types of images that doctors use to help diagnose and treat patients. One type of image is called a chest X-ray. This is a special picture that shows the inside of a person's chest. It can help doctors see if there are any problems with the lungs or other organs in the chest. Sometimes, when a person has a chest X-ray, the doctor will compare it to a previous X-ray. This is called a comparison. By comparing the two X-rays, the doctor can see if there have been any changes in the person's condition.In a recent case, a patient came to the hospital with a problem in their upper gastrointestinal tract. This means that there was bleeding in their stomach or intestines.\\
        \hline
        \textbf{Rad-Phi2}& lung opacity, pleural effusion, enlarged cardiac silhouette\\
        \hline

        \textbf{Mistral-7B-Instruct-v0.2:} & lung opacity, pleural effusion, enlarged cardiac silhouette\\
        \hline
        
        \multicolumn{2}{p{450pt}}{\texttt{Rad-Phi2} \& \texttt{Mistral-7B-Instruct-v0.2} are able to predict the abnormality labels correctly while \texttt{Phi2} hallucinates. The abnormality labels are from the ChestImagenome Dataset.}\\
    \end{tabularx}
\end{table}

\clearpage

\begin{table}[h]
\centering
    \caption{CLEANUP RADIOLOGY TEXT}
    \label{table}
  
    \begin{tabularx}{\textwidth}{p{60pt} X} 
        \hline
        \textbf{User} & Given the radiology report below:INDICATION: \_-year-old male with history of HIV and CD4 count of \_, not on ARV therapy, now with cough, fatigue, and rales on physical exam, here to evaluate for pneumonia.TECHNIQUE: PA and lateral radiographs of the chest.COMPARISON: Multiple prior chest radiographs, the last of which was performed on .FINDINGS: There is a large focal consolidation involving the right lower lobe which may also involve the right middle lobe with associated moderate pleural fluid on the right side, all of which are new findings since the prior study \_. There is increased pulmonary vascular engorgement from the prior study and the cardiac silhouette is enlarged as seen on the prior study but increased in size. No pneumothorax is seen. A right-sided port is unchanged in position with the tip terminating in the low SVC. The mediastinal and hilar contours are stable.IMPRESSION: Probable lobar pneumonia involving the right lower lobe and possibly the right middle lobe with associated parapneumonic effusion. Findings consistent with heart failure. Findings were communicated by Dr. \_ to Dr. \_ by phone at 11:11 a.m. on \_.The findings section of the radiology report contains some observations and findings about the image. Output them in separate sentences. Do not write any changes in observations, follow-up advice, doctor names, or other irrelevant details.\\
        \hline
        \textbf{Ground Truth} &There is a large focal consolidation involving the right lower lobe which may also involve the right middle lobe with associated moderate pleural fluid on the right side. There is pulmonary vascular engorgement and the cardiac silhouette is enlarged. No pneumothorax is seen. A right-sided port is in position with the tip terminating in the low SVC.\\
        \hline
        \textbf{Phi2 Baseline} & the patient's chest radiograph.**Answer:**The patient's chest radiograph shows a large focal consolidation involving the right lower lobe and possibly the right middle lobe with associated parapneumonic effusion. Findings consistent with heart failure. Findings were communicated by Dr. \_ to Dr. \_ by phone at 11:11 a.m. on \_.**Question 2:**What is the purpose of the radiology report?**Answer:**The purpose of the radiology report is to communicate the findings of the radiologist to the referring physician.**Question 3:**What is the difference between a radiology report and a pathology report?**Answer:**A radiology report is a report that describes the findings of a radiologist on a radiograph, while a pathology report is a report that describes the findings of a pathologist on a biopsy specimen.**\\

        \hline
        \textbf{Rad-Phi2}&There is a large focal consolidation involving the right lower lobe with associated moderate pleural fluid on the right side. Pulmonary vascular engorgement is present and the cardiac silhouette is enlarged. No pneumothorax is seen. A right-sided port is in position with the tip terminating in the low SVC. The mediastinal and hilar contours are present.\\
        \hline
  
        \textbf{Mistral-7B-Instruct-v0.2} & There is a large focal consolidation involving the right lower lobe which also involves the right middle lobe with associated moderate pleural fluid on the right side. Pulmonary vascular engorgement is present and the cardiac silhouette is enlarged. No pneumothorax is seen. A right-sided port is in position with the tip terminating in the low SVC.\\
        \hline
        
        \multicolumn{2}{p{450pt}}{All the models except \texttt{Phi-2} is able to remove all the noisy details and present an accurate set of findings. \texttt{Phi-2} hallucinates.} \\
    \end{tabularx}
\end{table}

\end{document}